\documentclass[twoside]{article}
\usepackage[accepted]{aistats2018}
\usepackage{cite}
\usepackage{algorithm,algorithmicx}
\usepackage{algpseudocode}
\usepackage[small,bf]{caption}
\usepackage[cmex10]{amsmath}
\usepackage{amssymb,latexsym,epsfig,subfigure,epic,amscd,mathrsfs,euscript,eufrak,bm}
\usepackage{color}
\usepackage{booktabs}
\usepackage{array}
\usepackage{comment}
\usepackage{epstopdf}
\usepackage{amsfonts}
\usepackage{amsopn}
\usepackage{graphicx}
\usepackage{url} 
\usepackage{empheq}
\usepackage[dvipsnames]{xcolor}

\usepackage[round]{natbib}

\newtheorem{mycor}{\bf{Corollary}}
\newtheorem{mythr}{\bf{Theorem}}

\newcommand{\argmin}{\operatornamewithlimits{arg\,min}}

\DeclareMathOperator*{\minimize}{\text{minimize}}

\DeclareMathOperator*{\st}{\text{subject to}}
\DeclareMathAlphabet\mathbfcal{OMS}{cmsy}{b}{n}
\newcommand{\Def}[0]{\mathrel{\mathop:}=}
\DeclarePairedDelimiterX{\inp}[2]{\langle}{\rangle}{#1, #2}

\usepackage{color}
 
\begin{document}

%
\runningtitle{ZOO-ADMM: Convergence Analysis and Applications}

%

\twocolumn[

\aistatstitle{Zeroth-Order Online  Alternating Direction Method of Multipliers: 
Convergence Analysis and Applications}

\aistatsauthor{Sijia Liu \And Jie Chen \And    Pin-Yu Chen \And   Alfred O. Hero}

\aistatsaddress{University of Michigan\\
IBM Research, Cambridge 
\And  Northwestern Polytechnical\\
University, China
\And IBM  Research,\\ Yorktown Heights
\And  University of Michigan\\
}

]

\begin{abstract}
In this paper,
we design and analyze a new zeroth-order  online  algorithm, namely, the zeroth-order  online alternating direction method of multipliers (ZOO-ADMM), which  enjoys dual advantages of being gradient-free operation and  employing the ADMM   to accommodate  complex structured regularizers. 
Compared to the first-order   gradient-based   online  algorithm,  we show that
  ZOO-ADMM  requires $\sqrt{m}$ times more iterations, leading to    a convergence rate of $O(\sqrt{m}/\sqrt{T})$, 
where $m$ is the number of  optimization variables, and $T$ is the number of iterations. To accelerate ZOO-ADMM,
 we propose two minibatch strategies: gradient  sample  averaging   and  observation  averaging, resulting in an improved convergence rate of $O(\sqrt{1+q^{-1}m}/\sqrt{T})$, where $q$ is the minibatch size. In addition to convergence analysis, we also demonstrate   ZOO-ADMM  to applications in    signal processing, statistics, and  machine learning. 
\end{abstract}

\section{Introduction}
Online convex optimization (OCO) performs 
sequential inference in   
a data-driven adaptive fashion,
and has found a wide range of
applications  \citep{hazan2016introduction,hosseini2016online,hallwillett15}.
In this paper, we focus on   regularized convex optimization   in the   OCO setting, where a cumulative  empirical loss is minimized together with   a fixed regularization term.
Regularized loss minimization is a common learning paradigm, which  has 
 been very effective in promotion of sparsity through   $\ell_1$  or mixed $\ell_1$/$\ell_2$  regularization \citep{bach2012optimization}, low-rank matrix completion via nuclear norm regularization \citep{candes2009exact}, graph signal recovery via graph Laplacian regularization \citep{chen2017bias}, and constrained optimization by imposing  indicator functions of constraint sets \citep{parikh2014proximal}. 
 
 Several OCO algorithms have been proposed  for regularized  optimization, 
 e.g., composite mirror descent, namely, proximal stochastic gradient descent \citep{duchi2010composite},  regularized dual averaging \citep{xiao2010dual},   and adaptive gradient descent   \citep{duchi2011adaptive}.
 However, the complexity of the aforementioned  algorithms   
is dominated by the computation of     the proximal operation with respect to the regularizers  \citep{parikh2014proximal}.
An alternative is to use  online alternating direction method of multipliers (O-ADMM) \citep{suzuki2013dual,ouyang2013stochastic,wang2013online}. 
Different from the  algorithms  in \citep{duchi2010composite,xiao2010dual,duchi2011adaptive}, 
the ADMM framework offers the possibility of splitting the
  optimization
problem into a sequence of  easily-solved subproblems.  
It was shown in \citep{suzuki2013dual,ouyang2013stochastic,wang2013online} that   the   online variant of ADMM 
has   convergence rate of 
$O(1/\sqrt{T})$ for
convex loss functions and $O(\log{T}/T)$ for strongly
convex loss functions, where $T$ is the number of iterations.

One limitation of 
  existing  O-ADMM algorithms
is  the need to compute and repeatedly evaluate   the gradient of the loss function over the iterations. In many    practical scenarios, an  explicit expression for the  gradient is difficult to obtain. For example, 
in bandit optimization \citep{agarwal2010optimal}, a player  receives  partial  feedback in  terms of    loss function values  revealed  by her adversary, and  
making it impossible to compute the gradient of the full loss function. 
In adversarial black-box machine learning models, only the function values (e.g., prediction results) are provided \citep{chen2017zoo}.
Moreover, in some high dimensional settings,   acquiring the gradient
  information may be difficult, e.g.,   involving  matrix inversion \citep{boyd2004convex}. 
 This motivates the development of    gradient-free (zeroth-order) optimization algorithms.
  
Zeroth-order optimization approximates the full gradient via 
  a randomized gradient estimate 
  \citep{nesterov2015random,ghadimi2013stochastic,duchi2015optimal,agarwal2010optimal,shamir2017optimal,hajinezhadzenith}.
For example, in \citep{agarwal2010optimal,shamir2017optimal}, zeroth-order  algorithms were developed for bandit convex optimization with    multi-point  bandit feedback. 
In \citep{nesterov2015random},  a
zeroth-order gradient descent algorithm  was proposed that has   $O(m/\sqrt{T})$  convergence rate, where $m$ is the number of  variables in the objective function. 
A similar convergence rate was found in     \citep{ghadimi2013stochastic} for nonconvex optimization.
This slowdown (proportional to the problem size   $m$) in  convergence rate was further improved to $O(\sqrt{m}/\sqrt{T})$   \citep{duchi2015optimal}, whose optimality was  proved  under the framework of    mirror descent algorithms.
\textcolor{black}{A more recent relevant paper is \citep{gao2014information}, where a variant of the ADMM algorithm that uses gradient estimation was introduced.  However, the ADMM algorithm presented  in \citep{gao2014information} was not customized for OCO. Furthermore, it   only ensured that the linear equality constraints are satisfied in expectation; hence, a particular instance of the proposed   solution could violate the   constraints. 
}


In this paper, we  propose a zeroth-order   online ADMM (called ZOO-ADMM) algorithm, and 
analyze 
its convergence rate
under different settings, including stochastic optimization, learning with strongly convex loss functions,  and      minibatch strategies for convergence acceleration.
 We summarize our contributions  as follows.
 
$\bullet$ We integrate the idea of zeroth-order optimization with online ADMM, leading to
 a new gradient-free OCO  algorithm, ZOO-ADMM.
 
$\bullet$ We prove ZOO-ADMM yields a $O(\sqrt{m}/\sqrt{T})$ 
     convergence rate for smooth+nonsmooth composite objective functions. 
     
$\bullet$ We introduce  a general hybrid minibatch strategy for acceleration of  ZOO-ADMM,
leading to an  improved convergence rate  $O(\sqrt{1+q^{-1}m}/\sqrt{T})$, where $q$ is the minibatch size.

$\bullet$ We illustrate the practical utility of  ZOO-ADMM in machine leanring, signal processing and statistics. 

\section{ADMM: from First  to Zeroth Order}
\label{sec: prob}
In this paper, we consider
the   regularized loss minimization   problem  over a time horizon of length $T$
\begin{align}
\begin{array}{ll}
    \displaystyle \minimize_{\mathbf x \in \mathcal X, \mathbf y \in \mathcal Y} &  \displaystyle  \frac{1}{T} \sum_{t=1}^T  f(\mathbf x; \mathbf w_t) 
    + \phi(\mathbf y) \\
   \st  & \mathbf A \mathbf x + \mathbf B \mathbf y = \mathbf c,
\end{array}
\label{eq: prob_online_reg_xy}
\end{align}
where $\mathbf x \in \mathbb R^m$ and $\mathbf y \in \mathbb R^d$ are optimization variables, $\mathcal X  $ and $\mathcal Y$ are closed convex sets, $f(\cdot; \mathbf w_t)$  is a convex and smooth cost/loss function parameterized by   $\mathbf w_t$ at time $t$, $\phi$ is a convex {regularization} function (possibly {nonsmooth}), and $\mathbf A \in \mathbb R^{l \times m}$,   $\mathbf B \in \mathbb R^{l \times d}$, and $\mathbf c \in \mathbb R^l$ are appropriate coefficients associated with a  system of $l$  linear constraints.

In problem \eqref{eq: prob_online_reg_xy},    
the use of time-varying cost functions $\{ f(\mathbf x; \mathbf w_t) \}_{t=1}^T$   captures possibly time-varying environmental uncertainties that may exist  in the online setting \citep{hazan2016introduction,shalev2012online}.  
We can also write the online cost as $f_t(\mathbf x)$   when it cannot  be explicitly parameterized by $\mathbf w_t$. 
One interpretation of $\{ f(\mathbf x; \mathbf w_t) \}_{t=1}^T$  is the empirical approximation to the   stochastic objective function
$\mathbb E_{\mathbf w\sim P}   \left [ f(\mathbf x; \mathbf w) \right ]$. 
Here
$P$ is an empirical distribution with density $\sum_t \delta(\mathbf w, \mathbf w_t)$, where $\{ \mathbf w_t \}_{t=1}^T$ is a set of i.i.d. samples, and $\delta(\cdot, \mathbf w_t)$ is the Dirac  delta  function at $\mathbf w_t$. 
We also note that when $\mathcal Y = \mathcal X$, $l=m$, $\mathbf A = \mathbf I_m$, $\mathbf B = -\mathbf I_m$, $\mathbf c = \mathbf 0_m$, 
the variable $\mathbf y$ and the linear constraint in
  \eqref{eq: prob_online_reg_xy} can be eliminated, leading to a standard OCO formulation. Here    $\mathbf I_m$ denotes the $m \times m$ identity matrix, and $\mathbf 0_m$ is the $m \times 1$ vector of all zeros\footnote{In the sequel we will omit the dimension index $m$, which can be inferred from the context.}.

\subsection{Background on O-ADMM}
O-ADMM   \citep{suzuki2013dual,wang2013online,ouyang2013stochastic} was originally proposed 
 to extend batch-type ADMM methods to the OCO setting. For
solving \eqref{eq: prob_online_reg_xy},
a widely-used  algorithm was developed by   \citep{suzuki2013dual}, which combines online   proximal gradient descent and   ADMM in the following form:
\begin{align}
& \mathbf x_{t+1} = \argmin_{\mathbf x \in \mathcal X} \left \{  \mathbf g_t^T \mathbf x - \boldsymbol \lambda_t^T (\mathbf A \mathbf x + \mathbf B \mathbf y_t - \mathbf c) \right. \nonumber \\
& \hspace*{0.3in} \left. + \frac{\rho}{2} \left \| \mathbf A \mathbf x + \mathbf B \mathbf y_t - \mathbf c \right \|_2^2 + \frac{1}{2 \eta_t}  \| \mathbf x - \mathbf x_t \|_{\mathbf G_t}^2 \right \}, \label{eq: x_step} \\
& \mathbf y_{t+1} = \argmin_{\mathbf y \in \mathcal Y} \left  \{  \phi(\mathbf y) -  \boldsymbol \lambda_t^T (\mathbf A \mathbf x_{t+1} + \mathbf B \mathbf y  - \mathbf c) \right. \nonumber \\
& \hspace*{0.3in} \left. + \frac{\rho}{2} \| \mathbf A \mathbf x_{t+1} + \mathbf B \mathbf y  - \mathbf c\|_2^2  \right \}, \label{eq: y_step} \\
& \boldsymbol \lambda_{t+1} =  \boldsymbol \lambda_t - \rho (\mathbf A \mathbf x_{t+1} + \mathbf B \mathbf y_{t+1}  - \mathbf c), \label{eq: dual_step}
\end{align}
where $t$ is the iteration number (possibly the same as the   time step), 
 $\mathbf g_t$ is  the gradient of the cost function $f(\mathbf x; \mathbf w_t)$  
 at $\mathbf x_t$, namely, $\mathbf g_t = \nabla_{\mathbf x} f(\mathbf x;\mathbf w_t) |_{\mathbf x = \mathbf x_t}$,
 $ \boldsymbol \lambda_t$ is a Lagrange multiplier (also known as the dual variable), $\rho$ is a positive   weight to penalize
 the augmented term associated with the equality constraint of~\eqref{eq: prob_online_reg_xy}, 
 $\| \cdot \|_2
 $ denotes the $\ell_2$ norm,
 $\eta_t$ is a non-increasing sequence of positive step sizes, and $\| \mathbf x - \mathbf x_t \|_{\mathbf G_t}^2 = (\mathbf x - \mathbf x_t)^T \mathbf G_t (\mathbf x - \mathbf x_t)$ is a Bregman divergence  generated by the strongly convex function $(1/2)\mathbf x^T \mathbf G_t \mathbf x$ with a known symmetric positive definite coefficient matrix $\mathbf G_t$. 
 
Similar to batch-type ADMM algorithms,   the subproblem in \eqref{eq: y_step}  is often easily solved via the proximal operator with respect to $\phi$ 
\citep{boyd2011distributed}. 
 However, one limitation of O-ADMM is that  it requires the gradient 
$\mathbf g_t$ in \eqref{eq: x_step}. 
We will develop the
gradient-free (zeroth-order)   O-ADMM algorithm  below that relaxes this requirement. 

\subsection{Motivation of ZOO-ADMM}
To avoid explicit gradient calculations in \eqref{eq: x_step}, we adopt a  
random gradient estimator to estimate the   gradient of a smooth cost function 
 \citep{nesterov2015random,ghadimi2013stochastic,duchi2015optimal,shamir2017optimal}.
 The    gradient estimate of $f(\mathbf w; \mathbf w_t)$ is given by
 \begin{align}
\hat {\mathbf g}_t = \frac{f(\mathbf x_t + \beta_t \mathbf z_t; \mathbf w_t) - f(\mathbf x_t; \mathbf w_t  ) }{\beta_t} \mathbf z_t, \label{eq: gt_0}
\end{align}
where $ \mathbf z_t \in \mathbb R^m $ is a random vector drawn  independently at each iteration $t$  from a distribution $\mathbf z \sim \mu$
with $\mathbb E_\mu[   \mathbf z  \mathbf z^T ] = \mathbf I$, 
and $\{  \beta_t \}$ is a non-increasing sequence of small positive smoothing constants. 
Here   for notational simplicity we replace $\{ \}_{t=1}^T$ with
 $\{ \}$.
The rationale behind the estimator \eqref{eq: gt_0} is that $\hat {\mathbf g}_t $ becomes an   unbiased estimator of $\mathbf g_t$ when the smoothing parameter $\beta_t$ approaches zero \citep{duchi2015optimal}.

After replacing  $\mathbf g_t$ with $\hat {\mathbf g}_t$ in \eqref{eq: gt_0}, the resulting algorithm \eqref{eq: x_step}-\eqref{eq: dual_step} can be implemented without explicit gradient computation.
This extension is called zeroth-order O-ADMM (ZOO-ADMM) that involves a modification of step
  \eqref{eq: x_step}  :
\begin{align}
 \mathbf x_{t+1} & =  \argmin_{\mathbf x \in \mathcal X} \left \{  \hat{\mathbf g}_t^T \mathbf x - \boldsymbol \lambda_t^T (\mathbf A \mathbf x + \mathbf B \mathbf y_t - \mathbf c) \right. \nonumber\\ 
 &    \left. + \frac{\rho}{2} \left \| \mathbf A \mathbf x + \mathbf B \mathbf y_t - \mathbf c \right \|_2^2 + \frac{1}{2 \eta_t}  \| \mathbf x - \mathbf x_t \|_{\mathbf G_t}^2 \right \}. \label{eq: x_step_ZOADMM}  
\end{align}
In 
 \eqref{eq: x_step_ZOADMM}, we can specify the matrix $\mathbf G_t$ in such a way as to cancel the term   $\| \mathbf A \mathbf x\|_2^2$. 
This technique   has been used in  the linearized ADMM algorithms \citep{parikh2014proximal,zhang2011unified} to avoid matrix inversions. Defining
$\mathbf G_t = \alpha \mathbf I - \rho \eta_t \mathbf A^T \mathbf A$, 
 the update rule \eqref{eq: x_step_ZOADMM} simplifies to  a projection operator
\begin{align}
 &\mathbf x_{t+1} = \argmin_{\mathbf x \in \mathcal X} \left \{   
\left \| \mathbf x -  \boldsymbol \omega  \right  \|_2^2 
 \right \} ~ \text{with}  \label{eq: x_step_simple}   \\
&  \boldsymbol \omega \Def 
\left [ 
 \frac{\eta_t}{\alpha} \left (
- \hat {\mathbf g}_t + \mathbf A^T ( \boldsymbol \lambda_t - \rho ( \mathbf A \mathbf x_t +  \mathbf B \mathbf y_t -  \mathbf c) )
\right )+ \mathbf x_t 
\right ], \nonumber 
\end{align}
where 
$\alpha >0$ is a  parameter selected to ensure  $\mathbf G_t \succeq \mathbf I$. Here  $\mathbf X \succeq \mathbf Y$ signifies that $\mathbf X - \mathbf Y$ is positive semidefinite.

To evaluate the convergence behavior of ZOO-ADMM, we will derive   its expected average regret \citep{hazan2016introduction}   
\begin{align}
\overline{\mathrm{Regret}}_{T}(  \mathbf x_t, \mathbf y_t, \mathbf x^*, \mathbf y^* ) \Def & \mathbb E \left [ \frac{1}{T} \sum_{t=1}^T \left (
 f(\mathbf x_t;\mathbf w_t) + \phi(\mathbf y_t)
 \right ) \right. \nonumber \\
 & \hspace*{-0.6in}\left. - \frac{1}{T} \sum_{t=1}^T  \left (
 f(\mathbf x^*;\mathbf w_t) + \phi(\mathbf y^*)
 \right )\right ], \label{eq: ave_reg}
\end{align}
where $(\mathbf x^*, \mathbf y^*)$ denotes the best batch offline  solution.
 
\section{Algorithm and Convergence Analysis of ZOO-ADMM}
In this section, we begin by stating assumptions
 used in our analysis. We then formally define the ZOO-ADMM algorithm  and derive  its convergence rate.  
 
We assume the following conditions in our analysis.
\\
$\bullet$ \textit{Assumption A:}  In problem \eqref{eq: prob_online_reg_xy}, $\mathcal X$ and $\mathcal Y$ 
are bounded with
 finite diameter $R$, 
and at least one of $\mathbf A$ and $\mathbf B$ in $\mathbf A \mathbf x + \mathbf B \mathbf y = \mathbf c$ is invertible. 

$\bullet$ \textit{Assumption B:}   $f(\cdot;\mathbf w_t)$ is convex and Lipschitz continuous with  $\sqrt{\mathbb E [ \| \nabla_{\mathbf x} f(\mathbf x;\mathbf w_t) \|_2^2 ]} \leq L_1$ for all  $t$ and $\mathbf x \in \mathcal X$.

$\bullet$ \text{Assumption C:}  $f(\cdot; \mathbf w_t)$ is $L_{g}(\mathbf w_t)$-smooth with
 $L_g = \sqrt{\mathbb E[(L_g(\mathbf w_t)^2)]}$.

$\bullet$ \textit{Assumption D:}  
 $\phi$ is convex and  $L_2$-Lipschitz continuous with $ \| \partial \phi(\mathbf y) \|_2  \leq L_2$ for all $\mathbf y \in \mathcal Y$, where $\partial \phi(\mathbf y)$ denotes the subgradient of $\phi$. 
 
 
$\bullet$  \textit{Assumption E:} In \eqref{eq: gt_0}, given  $\mathbf z \sim \mu$, the quantity $M(\mu) \Def \sqrt{\mathbb E [\| \mathbf z \|_2^6 ]}$ is finite, and there is a function $s : \, \mathbb N \to \mathbb R_+$ satisfying
 $
 \mathbb E[\|\langle \mathbf a, \mathbf z \rangle \mathbf z \|_2^2] \leq s(m) \|  \mathbf a \|_2^2
 $ for all $\mathbf a \in \mathbb R^m$,   
 where  $\langle \cdot, \cdot \rangle $ denotes the inner product of two vectors.

We remark that Assumptions A-D  are standard for  stochastic gradient-based and ADMM-type
 methods \citep{hazan2016introduction,shalev2012online,boyd2011distributed,suzuki2013dual}. We elaborate on the rationale behind them in Sec.\,\ref{sec: assump}.
Assumption E
places moment constraints on the distribution $\mu$ that will allow us to derive the necessary concentration bounds for our convergence analysis.
If 
  $\mu$ is uniform on the surface of the Euclidean-ball of
radius $\sqrt{m}$, we  have $M(\mu)  = m^{1.5} $ and $s(m) =  m $. And if $\mu = \mathcal N(\mathbf 0, \mathbf I_{m \times m})$, we have  $M(\mu)  \approx m^{1.5} $ and $s(m) \approx  m $ \citep{duchi2015optimal}. 
For ease of representation,  we restrict our attention to the case that $s(m) = m$ in the rest of the paper. 
It is also worth mentioning that the convex and strongly convex conditions of   $f(\cdot;\mathbf w_t)$ can be described as 
\begin{align}
f(\mathbf x;\mathbf w_t) \geq & f(\tilde {\mathbf x};\mathbf w_t) + (\mathbf x - \tilde {\mathbf x})^T \nabla_{\mathbf x} f(\tilde {\mathbf x};\mathbf w_t) \nonumber \\
& + \frac{\sigma}{2} \| \mathbf x - \tilde {\mathbf x} \|^2, ~ \forall \mathbf x, \tilde {\mathbf x}, \label{eq: strong_convex}
\end{align}
where $\sigma \geq 0$ is   a   parameter  controlling convexity. If 
$\sigma > 0$, then $f(\cdot; \mathbf w_t)$ is strongly convex with parameter $\sigma$. Otherwise ($\sigma = 0$), \eqref{eq: strong_convex}  implies  convexity of $f(\cdot; \mathbf w_t)$.

The
    ZOO-ADMM     iterations are given as Algorithm\,1. 
     Compared to O-ADMM in \citep{suzuki2013dual},
      we only require querying two function values for the generation of gradient estimate  at step\,3.
Also different from \citep{gao2014information}, steps 7-11 of Algorithm\,1 imply that
the equality constraint of problem \eqref{eq: prob_online_reg_xy} is always satisfied at $\{ \mathbf x_t, \mathbf y_t^\prime \}$ or $\{ \mathbf x_t^\prime, \mathbf y_t \}$.
The      average regret of ZOO-ADMM  is bounded in Theorem\,\ref{prop: ZO_regret}.

 \begin{mythr}\label{prop: ZO_regret}
Suppose $\mathbf B$ is invertible in problem \eqref{eq: prob_online_reg_xy}. For
$\{ \mathbf x_t, \mathbf y_t^\prime \}$ generated by ZOO-ADMM, the expected average regret is bounded as
\begin{align}\label{eq: regret_ZOADMM_general}
 & \overline{\mathrm{Regret}}_{T}(  \mathbf x_t, \mathbf y_t^\prime, \mathbf x^*, \mathbf y^* )
\nonumber \\
  \leq &
\frac{1}{T} \sum_{t=2}^T \max\{  \frac{\alpha}{2\eta_t} - \frac{\alpha}{2\eta_{t-1}}  - \frac{\sigma}{2} , 0 \} R^2 +     \frac{m L_1^2}{T}\sum_{t=1}^T \eta_t  \nonumber \\
&+ \frac{M(\mu)^2 L_g^2}{4T}\sum_{t=1}^T \eta_t \beta_t^2 + \frac{K}{T},
\end{align}
where 
$\alpha$ is introduced in \eqref{eq: x_step_simple}, $R$, $L_1$, $L_g$,   $s(m)$ and  $M(\mu)$  are defined in Assumptions A-E, and $K$   denotes a constant term   that depends on  $\alpha$, $R$, $\eta_1$, $\mathbf A$, $\mathbf B$, $\boldsymbol \lambda$, $\rho$ and $L_2$.
Suppose $\mathbf A$ is invertible in problem \eqref{eq: prob_online_reg_xy}. 
For   $\{ \mathbf x_t^\prime, \mathbf y_t \}$, 
the regret $\overline{\mathrm{Regret}}_{T}(  \mathbf x_t^\prime, \mathbf y_t, \mathbf x^*, \mathbf y^* )$ obeys the same bounds as 
\eqref{eq: regret_ZOADMM_general}. 
\end{mythr}
\textbf{Proof:} See Sec.\,\ref{sec: thr1}.
\hfill $\blacksquare$

In Theorem\,\ref{prop: ZO_regret}, if the step size $\eta_t$ and the smoothing parameter $\beta_t$  are chosen as
\begin{align}\label{eq: step_size_specific}
\eta_t = \frac{C_1}{ m \sqrt{t}},~ \beta_t = \frac{C_2}{{M(\mu)}t}
\end{align}
for some constant $C_1>0$ and $C_2 > 0$,    then the  regret bound \eqref{eq: regret_ZOADMM_general}  simplifies to
\begin{align}\label{eq: regret_ZOADMM}
& \overline{\mathrm{Regret}}_{T}(  \mathbf x_t, \mathbf y_t^\prime, \mathbf x^*, \mathbf y^* )  
  \leq  
\frac{\alpha R^2 }{2C_1}  \frac{\sqrt{m}}{\sqrt{T}}  \nonumber \\
& \hspace*{0.5in}+    2 C_1  L_1^2  \frac{\sqrt{m}}{\sqrt{T}}  
+ \frac{5C_1 C_2^2 L_g^2}{12} \frac{1}{T}
+ \frac{K}{T}.
\end{align}
The above simplification is derived in Sec.\,\ref{sec: simplify}. 

\begin{algorithm}
\caption{ZOO-ADMM for solving problem  \eqref{eq: prob_online_reg_xy}}
\begin{algorithmic}[1]
\State Input: $\mathbf x_1 \in \mathcal X$, $\mathbf y_1 \in \mathcal Y$,  $\boldsymbol \lambda_1 = \mathbf 0$, $\rho >0$, 
step sizes $\{ \eta_t \}$,   smoothing constants $\{ \beta_t \}$,  distribution $\mu$, 
  and $\alpha \geq  \rho \eta_t \lambda_{\max}( \mathbf A^T \mathbf A  ) + 1$ so that
$\mathbf G_t \succeq \mathbf I$, where $\lambda_{\max}(\cdot)$ denotes the maximum eigenvalue of a symmetric matrix
\For{$t =  1,2,\ldots, T$}
\State sample $\mathbf z_t \sim \mu$  to generate    $\hat{\mathbf g}_t$ using \eqref{eq: gt_0}
\State update $\mathbf x_{t+1}$ via 
\eqref{eq: x_step_simple}  under  $\hat{\mathbf g}_t$ and $(\mathbf x_t, \mathbf y_t, \boldsymbol \lambda_t)$
\State update $\mathbf y_{t+1}$ via \eqref{eq: y_step}  under $(\mathbf x_{t+1},  \boldsymbol \lambda_t)$
\State update $\boldsymbol \lambda_{t+1}$ via \eqref{eq: dual_step}  under $(\mathbf x_{t+1}, \mathbf y_{t+1}, \boldsymbol \lambda_t)$
 \If  {$\mathbf B$ is invertible}  
\State compute $\mathbf y_{t+1}^\prime \Def \mathbf B^{-1}(\mathbf c - \mathbf A \mathbf x_{t+1})$ 
 \Else 
\State compute $\mathbf x_{t+1}^\prime \Def  \mathbf A^{-1}(\mathbf c - \mathbf B \mathbf y_{t+1})$ 
 \EndIf
\EndFor  
\State output: $\{ \mathbf x_t, \mathbf y_t^\prime \}$  or $\{ \mathbf x_t^\prime, \mathbf y_t \}$, 
running average $( \bar{\mathbf x}_{T} , \bar{\mathbf y}_{T}^\prime  )$ or
$(\bar{\mathbf x}_{T}^\prime , \bar{\mathbf y}_{T}  )$, where $\bar{\mathbf x}_{T} = \frac{1}{T} \sum_{k=1}^{T} \mathbf x_{k}$. 
\end{algorithmic}
\end{algorithm}

 It is clear from  \eqref{eq: regret_ZOADMM} that ZOO-ADMM converges at least as fast as 
 $O(\sqrt{m}/\sqrt{T})$, which is similar to the convergence rate of O-ADMM found by \citep{suzuki2013dual} but involves an additional factor   $\sqrt{m}$. 
 Such a dimension-dependent effect on the convergence rate has also been reported for other zeroth-order optimization algorithms \citep{ghadimi2013stochastic,duchi2015optimal,shamir2017optimal}, leading to the same convergence rate as ours.
In \eqref{eq: regret_ZOADMM},  even if we set $C_2 = 0$ (namely, $\beta_t = 0$) for an unbiased gradient estimate   \eqref{eq: gt_0}, the dimension-dependent factor $\sqrt{m}$ is not eliminated. That is because
the   second moment of the gradient estimate   also depends  
  on the number of  optimization variables. 
In the next section, we will propose    
 two minibatch strategies that can be used to reduce the variance of the gradient estimate  and to improve the convergence speed of   ZOO-ADMM.

\section{Convergence for Special Cases}
In this section,  
we specialize  ZOO-ADMM to three cases: 
a) stochastic optimization, 
b)  strongly convex cost  function  in    \eqref{eq: prob_online_reg_xy},
and c) the use of minibatch strategies for evaluation of gradient estimates.
Without loss of generality, we restrict analysis to the case that $\mathbf B$ is invertible in   \eqref{eq: prob_online_reg_xy}.

The  stochastic optimization  problem
is a special case of the OCO problem \eqref{eq: prob_online_reg_xy}.
If the objective function becomes  $F(\mathbf x, \mathbf y) \Def  \mathbb E_{\mathbf w} [f(\mathbf x; \mathbf w)] + \phi(\mathbf y)$ then  
 we can   link the regret with the optimization error at  the    running average $\bar{\mathbf x}_T$ and $\bar{\mathbf y}_T$ under the condition that $F$ is convex.
We state our results  as Corollary\,\ref{col: error_stoch}.
\begin{mycor}\label{col: error_stoch}
 Consider the stochastic optimization problem with the objective function $F(\mathbf x, \mathbf y) \Def  \mathbb E_{\mathbf w} [f(\mathbf x; \mathbf w)] + \phi(\mathbf y)$,
 and set
$\eta_t$ and 
$\beta_t$ using \eqref{eq: step_size_specific}.  
For
$\{ \bar{\mathbf x}_t, \bar{\mathbf y}_t^\prime\}$ generated by ZOO-ADMM, the optimization error
$\mathbb E \left [  F(\bar{\mathbf x}_T, \bar{\mathbf y}_T^\prime) -  F(\mathbf x^*, \mathbf y^*) \right ]$
 obeys the same bound as \eqref{eq: regret_ZOADMM}.
\end{mycor}
\textbf{Proof:} See Sec.\,\ref{sec: coro1}.
 \hfill $\blacksquare$
 
We recall from \eqref{eq: strong_convex} that $\sigma$ controls the convexity    of $f_t$, where $\sigma > 0$ if $f_t$ is strongly convex. In Corollary\,\ref{col: strong_convex}, we show that   $\sigma  $ 
 affects   
 the average regret  of ZOO-ADMM.
 
\begin{mycor}\label{col: strong_convex}
Suppose $f(\cdot; \mathbf w_t)$ is strongly convex, and  the step size $\eta_t$ and the smoothing parameter $\beta_t$ are chosen as
$\eta_t = \frac{\alpha}{ \sigma  {t}}$ and 
$\beta_t = \frac{C_2}{{M(\mu)}t}$ for    $C_2 > 0$.  Given
$\{ \mathbf x_t, \mathbf y_t^\prime \}$ generated by ZOO-ADMM, the expected average regret can be bounded as
\begin{align}\label{eq: regre_strong}
\overline{\mathrm{Regret}}_{T}(  \mathbf x_t, \mathbf y_t^\prime, \mathbf x^*, \mathbf y^* )  \leq &
     \frac{\alpha  L_1^2}{\sigma } \frac{m \log{T}}{T} \nonumber \\
     & +\frac{ 3\alpha C_2^2 L_g^2}{8 \sigma} \frac{1}{T} + \frac{K}{T}.
\end{align}
\end{mycor}
\textbf{Proof:} See Sec.\,\ref{sec: coro2}.
\hfill $\blacksquare$

   Corollary\,\ref{col: strong_convex} implies that  when the cost function is strongly convex, 
  the   regret bound  of ZOO-ADMM could achieve
  $O(m/{{T}})$ up to a logarithmic factor $\log{T}$. 
Compared to the regret bound $O({\sqrt{m}}/{\sqrt{T}})$ in the general case \eqref{eq: regret_ZOADMM}, the condition of strong convexity improves the regret bound in  terms of  the number of iterations $T$, but the dimension-dependent factor  now becomes linear in the dimension $m$ due to the effect of the second moment of gradient estimate.

The use of a  gradient estimator   makes the convergence  rate of ZOO-ADMM dependent on
the dimension $m$, i.e., the
number of optimization variables. Thus, it is important to study the impact of  minibatch strategies on the acceleration of the convergence speed \citep{li2014efficient,cotter2011better,suzuki2013dual,duchi2015optimal}. 
Here we present two   minibatch strategies:
{gradient sample  averaging and observation  averaging}.
In the first strategy, 
instead of using a single sample as in \eqref{eq: gt_0}, the average of $q$  sub-samples
 $\{ \mathbf z_{t,i} \}_{i=1}^q$ are used for gradient estimation
 \begin{align}
\hat {\mathbf g}_t 
= \frac{1}{q}\sum_{i=1}^q
\frac{f(\mathbf x_t + \beta_t \mathbf z_{t,i}; \mathbf w_t) - f(\mathbf x_t ; \mathbf w_t) }{\beta_t} \mathbf z_{t,i},
\label{eq: grad_mini_z}
\end{align} 
where $q$ is called the batch size. The use of \eqref{eq: grad_mini_z} is analogous to the use of an average gradient in incremental gradient \citep{blatt2007convergent} and stochastic gradient \citep{roux2012stochastic}.
  In the second strategy, 
   we     use a subset of observations $\{ \mathbf w_{t,i} \}_{i=1}^q$  to reduce the gradient variance,  
    \begin{align}
\hat {\mathbf g}_t 
= \frac{1}{q}\sum_{i=1}^q
\frac{f(\mathbf x_t + \beta_t \mathbf z_{t} ; \mathbf w_{t,i}) - f(\mathbf x_t  ; \mathbf w_{t,i}) }{\beta_t} \mathbf z_{t}.
\label{eq: grad_mini_w}
\end{align} 
\textcolor{black}
{We note that
in the online setting, the subset of observations $\{ \mathbf w_{t,i} \}_{i=1}^q$ can be obtained via a sliding time window of length $q$, namely, $\mathbf w_{i,t} = \mathbf w_{t-i+1}$ for $i = 1,2,\ldots, q$.
}


Combination of  \eqref{eq: grad_mini_z} and \eqref{eq: grad_mini_w}  yields a hybrid  strategy
\begin{align}
\hat {\mathbf g}_t =\frac{1}{q_1 q_2}\sum_{j=1}^{q_1}  \sum_{i=1}^{q_2} 
\frac{f(\mathbf x_t + \beta_t \mathbf z_{t,j} ; \mathbf w_{t,i}) - f(\mathbf x_t  ; \mathbf w_{t,i}) }{\beta_t} \mathbf z_{t,j}. \label{eq: grad_mini_wz}
\end{align}
In  Corollary\,\ref{col: mini_batch}, we demonstrate the convergence behavior of  the general hybrid  ZOO-ADMM. 

\begin{mycor}\label{col: mini_batch}
Consider the hybrid   minibatch strategy \eqref{eq: grad_mini_wz} in ZOO-ADMM, and set
$\eta_t = \frac{C_1}{ \sqrt{1+\frac{m}{q_1 q_2}} \sqrt{t}}$ and 
$\beta_t = \frac{C_2}{{M(\mu)}t}$. The expected average regret is bounded as
\begin{align}
&\overline{\mathrm{Regret}}_{T}(  \mathbf x_t, \mathbf y_t^\prime, \mathbf x^*, \mathbf y^* ) 
\leq  
\frac{\alpha R^2 }{2C_1} \frac{\sqrt{1+\frac{s(m)}{q_1 q_2}}}{\sqrt{T}} \nonumber \\
&+ 2 C_1 L_1^2 \frac{\sqrt{1+\frac{s(m)}{q_1 q_2}}}{\sqrt{T}} +  \frac{5C_1 C_2^2 L_g^2}{6}  \frac{1}{T} + \frac{K}{T},
\label{eq: regret_minbatch_zw}
\end{align}
where $q_1$ and $q_2$ are number of  sub-samples $\{ \mathbf z_{t,i} \}$  and  $\{ \mathbf w_{t,i} \}$, respectively. 
\end{mycor}
\textbf{Proof:} See Sec.\,\ref{sec: coro3}. \hfill $\blacksquare$

It is clear from  Corollary\,\ref{col: mini_batch}  that the use of minibatch strategies can alleviate the   dimension dependency, leading to the regret bound $O(\sqrt{1+m/(q_1 q_2)}/\sqrt{T})$. The regret bound in \eqref{eq: regret_minbatch_zw} also implies that   the convergence behavior of ZOO-ADMM 
  is similar using
  {either gradient sample averaging minibatch \eqref{eq: grad_mini_z} or observation averaging minibatch \eqref{eq: grad_mini_w}. If $q_1 = 1$ and $q_2 = 1$, the regret bound \eqref{eq: regret_minbatch_zw} reduces to
$O({\sqrt{m}}/{\sqrt{T}})$, which is the general case in \eqref{eq: regret_ZOADMM}.
If   $q_1q_2 = O(m)$, we obtain the   regret error $O({1}/{\sqrt{T}})$ as in the case where an explicit expression for the gradient is used in the 
OCO algorithms.}

\section{Applications of ZOO-ADMM}
\label{sec: app}
In this section, we demonstrate several applications of ZOO-ADMM   in  signal processing,  statistics and machine learning.   

\subsection{Black-box optimization}
\label{sec: black}
In some OCO problems, explicit gradient calculation  is impossible  due to the lack of 
a mathematical expression for   the loss function. 
For example,  commercial recommender systems try to build a representation of a customer's buying preference function based on a discrete number of queries or purchasing history, and the system never has access to the gradient of the user's preference function over their product line, which may even be unknown to the user. Gradient-free methods are therefore necessary. 
A specific example is the Yahoo! music recommendation system  \citep{dror2012yahoo}, which will be further discussed in the Sec.\,\ref{sec: exp}. 
In these examples, one can consider each user as a black-box model that provides feedback on the value of an objective function, e.g.,  relative preferences over all products, based on an online evaluation of the objective function at discrete points on its domain. Such a system can benefit from ZOO-ADMM.

\subsection{Sensor selection}
 \label{sec: sensrsel}
Sensor selection for parameter estimation is a fundamental problem   in  smart grids, communication systems, and wireless sensor networks \citep{hero2011sensor,liu2016sensor}. The goal is to seek the optimal  tradeoff between   sensor activations and the estimation accuracy. The  sensor selection problem is also closely related to  leader selection \citep{lin2014algorithms} and experimental design \citep{boyd2004convex}. 

For sensor selection, we often solve a (relaxed) convex program of the form  \citep{joshi2009sensor}
\begin{align}
\hspace*{-0.7in}\begin{array}{ll}
\displaystyle \minimize_{\mathbf x} & \displaystyle  \frac{1}{T}\sum_{t=1}^T \left [ - \mathrm{logdet} \left (\sum_{i=1}^m  x_i \mathbf a_{i,t} \mathbf a_{i,t}^T \right ) \right ] \vspace*{0.05in} \\
\st & \mathbf 1^T \mathbf x = m_0, ~ \mathbf 0 \leq \mathbf x \leq 1,
\end{array}
\hspace*{-0.5in}
\label{eq: prob_sensorSel_online}
\end{align}
 where $\mathbf x \in \mathbb R^m$ is the optimization variable,
 $m$ is the number of sensors, 
 $\mathbf a_{i,t} \in \mathbb R^n$ is the observation coefficient of sensor $i$ at time $t$, 
 and $m_0$ is the  number of selected sensors.  
The objective function of \eqref{eq: prob_sensorSel_online} can be interpreted as 
 the  log determinant of error covariance associated with the maximum likelihood estimator for parameter estimation \citep{rao1973linear}. 
The constraint $\mathbf 0 \leq \mathbf x \leq  \mathbf 1$ is a relaxed convex hull of the
 Boolean constraint   $\mathbf x \in \{0,1 \}^m$, which   encodes whether or not a sensor is selected.

Conventional   methods such as
  projected gradient (first-order)  and     interior-point (second-order) algorithms can be used to solve problem  \eqref{eq: prob_sensorSel_online}. However, both of them  involve  calculation of inverse matrices 
  necessary to evaluate
the gradient of  the   cost function. By contrast, 
  we can rewrite  \eqref{eq: prob_sensorSel_online} in a   form   amenable to  ZOO-ADMM  that avoids  matrix inversion,
\begin{align}
\begin{array}{ll}
\displaystyle \minimize_{\mathbf x,  \mathbf y} & \displaystyle \frac{1}{T} \sum_{t=1}^T f(\mathbf x; \mathbf w_t) + \mathcal I_1(\mathbf x) + \mathcal I_2(\mathbf y) \vspace*{0.05in} \\
\st &  \mathbf x - \mathbf y = \mathbf 0,
\end{array}
\label{eq: prob_sensorSel_online_ad}
\end{align}
where $\mathbf y \in \mathbb R^m$ is an auxiliary   variable,   $f(\mathbf x; \mathbf w_t) = - \mathrm{logdet}  (\sum_{i=1}^m  x_i \mathbf a_{i,t} \mathbf a_{i,t}^T  ) $ with $\mathbf w_t = \{\mathbf a_{i,t}\}_{i=1}^m$,  and  $\{ \mathcal I_i \}$ are indicator functions  
 \begin{align*}
\mathcal I_1(\mathbf x) = \left \{
\begin{array}{ll}
0 & \mathbf 0 \leq \mathbf x \leq \mathbf 1\\
\infty & \text{otherwise},
\end{array}
\right.
\mathcal I_2(\mathbf y) = \left \{
\begin{array}{ll}
0 & \mathbf 1^T \mathbf y = m_0\\
\infty & \text{otherwise}.
\end{array}
\right.
\end{align*}
We specify the ZOO-ADMM algorithm for solving \eqref{eq: prob_sensorSel_online_ad} in Sec.\,\ref{sec: sel_sol}. 

\subsection{Sparse Cox regression}
\label{sec: cox}

In survival analysis, Cox regression (also known as proportional hazards regression) is a  method to investigate 
  effects of   variables of interest    upon the amount of  time that elapses before a specified event  occurs, e.g., relating gene expression
profiles to survival time (time to cancer recurrence or
death) \citep{sohn2009gradient}.
Let $\{ \mathbf a_i \in \mathbb R^m,  \delta_i \in \{0,1\},t_i \in \mathbb R_+ \}_{i=1}^n$ be $n$ triples of $m$ covariates, where $\mathbf a_i$ is a   vector of covariates or factors for subject $i$, $\delta_i$ is a censoring indicator variable taking $1$ if
an event (e.g., death) is observed and $0$ otherwise, and $ t_i$ denotes the   censoring   time. 

This sparse regression problem can be formulated as the solution to an  $\ell_1$ penalized optimization  
 problem \citep{park2007l1,sohn2009gradient}, which yields
 \begin{align}
\displaystyle \minimize_{ {\mathbf x}} \quad  &\displaystyle \frac{1}{n}\sum_{i=1}^n \delta_i \left \{ - \mathbf a_i^T  {\mathbf x} +  \log{ \left (\sum_{j \in \mathcal R_i} e^{\mathbf a_j^T  {\mathbf x}} \right )}
\right \} \nonumber \\
& + \gamma \| {\mathbf x} \|_1 
\label{eq: cox_reg_ori}
\end{align}
where $ {\mathbf x} \in \mathbb R^m$ is the vector of covariates  coefficients to be designed, 
$\mathcal R_i$  is the   set of subjects at risk  at   time $t_i$, namely, $ \mathcal R_i = \{ j: t_j \geq t_i \}$, and $\gamma > 0$ is a regularization parameter.
In the objective function   of  \eqref{eq: cox_reg_ori}, the first term corresponds to 
the (negative)  log partial likelihood for the Cox proportional
hazards model \citep{cox1972}, and the second term encourages   sparsity of the covariate  coefficients.

By introducing a new variable $\mathbf y \in \mathbb R^m$ together with the constraint $\mathbf x - \mathbf y = \mathbf 0$,  problem \eqref{eq: cox_reg_ori} can be cast as the   canonical form  \eqref{eq: prob_online_reg_xy} amenable to the ZOO-ADMM algorithm. This helps us to avoid  the gradient calculation for the involved objective function in Cox regression. We specify the ZOO-ADMM algorithm for solving \eqref{eq: cox_reg_ori} in Sec.\,\ref{sec: cox_sol}. 

\section{Experiments} \label{sec: exp}
 In this section,  we demonstrate the effectiveness of ZOO-ADMM, and validate
its convergence behavior for the  applications  
introduced in 
 Sec.\,\ref{sec: app}.  In Algorithm\,1, 
  we set $\mathbf x_1 = \mathbf 0$, $\mathbf y_1 = \mathbf 0$, $\boldsymbol \lambda_1 = \mathbf 0$, $\rho = 10$, $\eta_t = 1/\sqrt{m t}$, $\beta_t = {1}/{(m^{1.5}t)}$,   $\alpha = \rho \eta_t \lambda_{\max}( \mathbf A^T \mathbf A  ) + 1$, and  the distribution
  $\mu$ is chosen to be uniform on the surface of the Euclidean-ball of
radius $\sqrt{m}$.
Unless specified otherwise, 
 we use the gradient sample averaging minibatch  of size $30$ in ZOO-ADMM.
Through this section, we  compare  ZOO-ADMM with the conventional O-ADMM algorithm in \citep{suzuki2013dual} under the same parameter settings.
Our experiments
are performed on a synthetic dataset
for sensor selection, 
and on real datasets for
 black-box optimization and    Cox regression.
Experiments were conducted by Matlab R2016 on a
 machine with 3.20 GHz CPU and 8 GB RAM.
 
    \begin{figure}[htb] 
  \centering
 \begin{tabular}{c}
\hspace*{-0.2in}\includegraphics[width=0.4\textwidth,height=!]{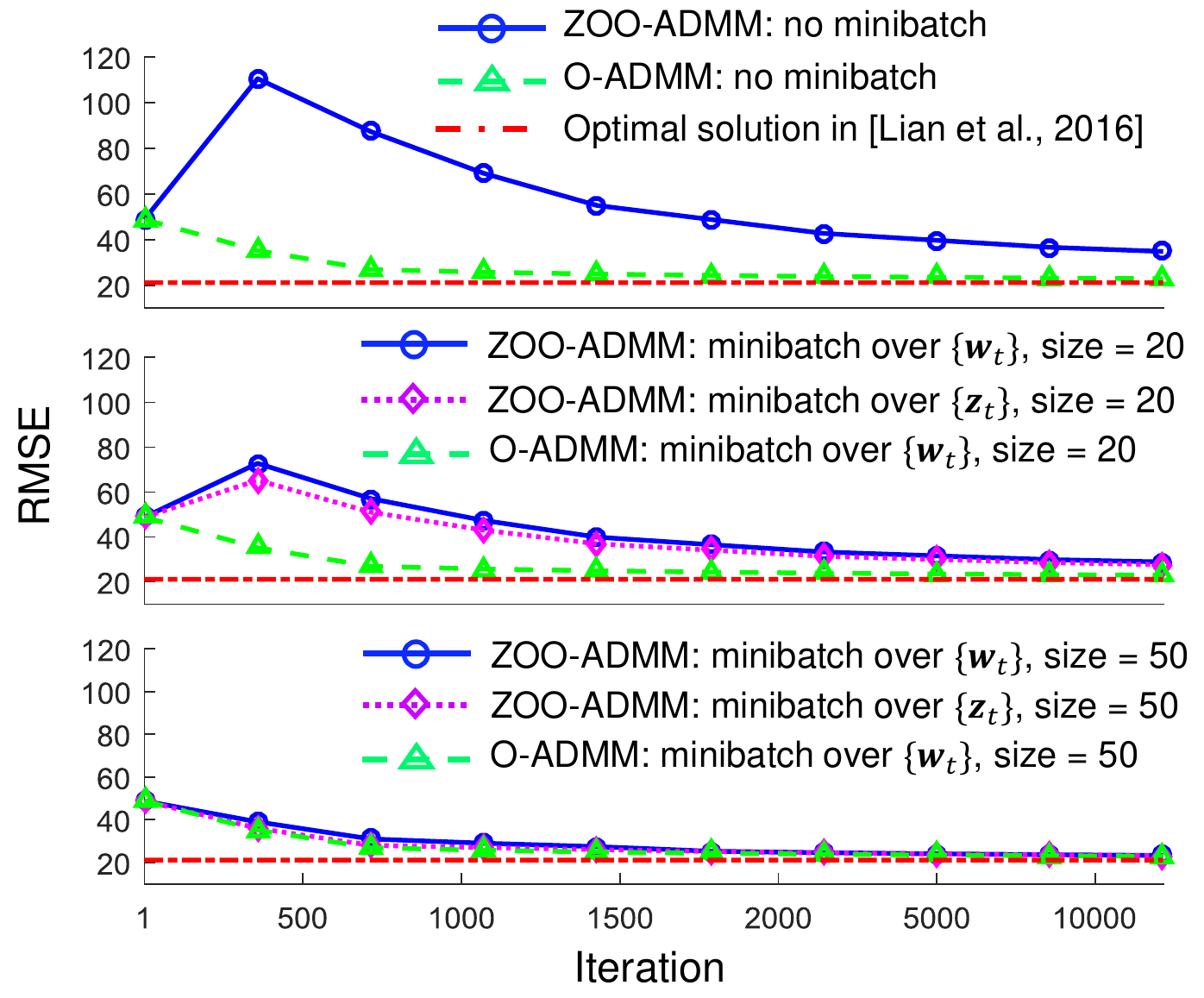} 
\\
(a) \\
\hspace*{-0.1in}\includegraphics[width=0.4\textwidth,height=!]{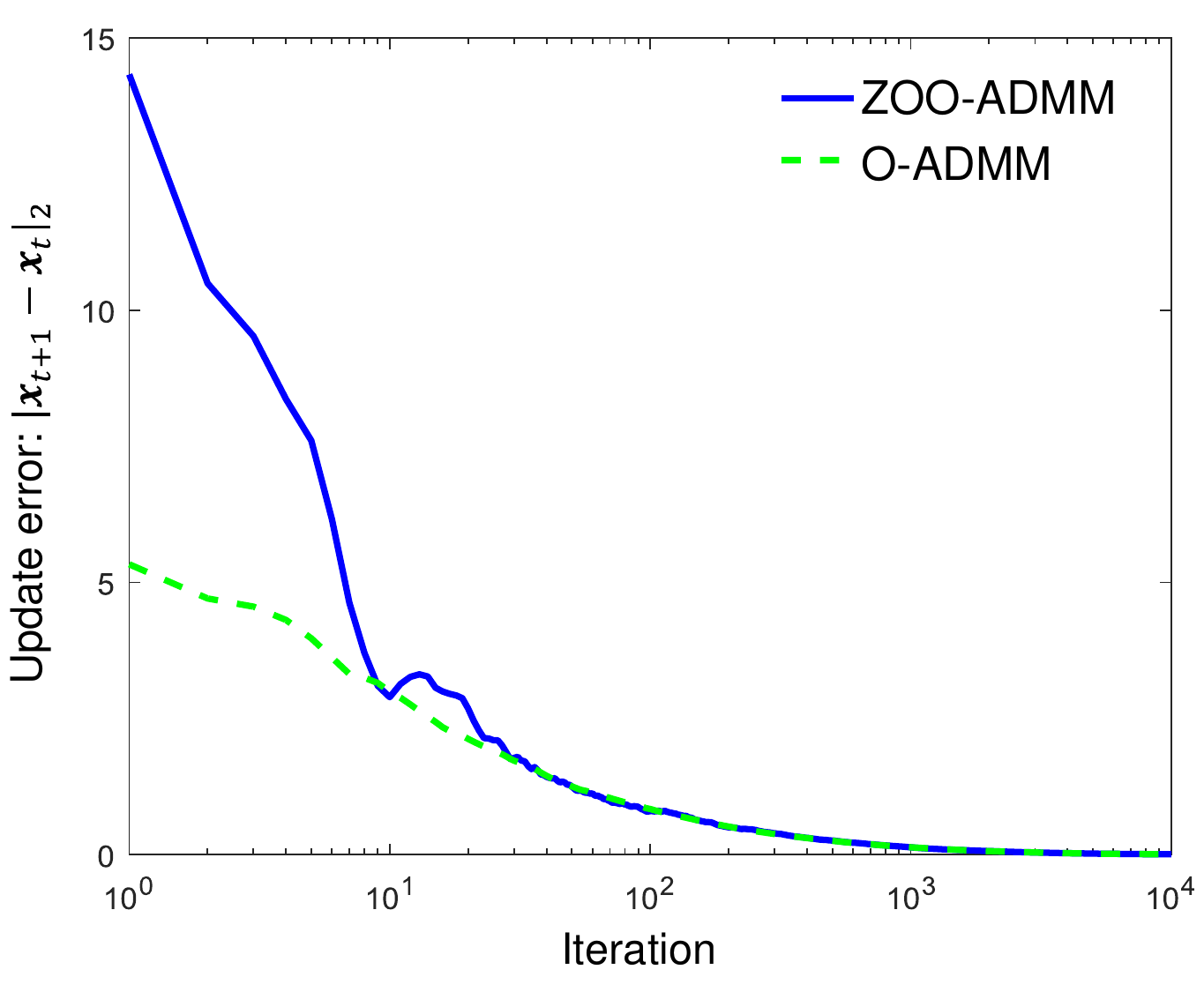}\\
  (b)
\end{tabular}
 \caption{Convergence of ZOO-ADMM: a) RMSE under different minibatch strategies, b) update error with minibatch size equal to $50$.}
\label{fig: black}
\end{figure}

 \textbf{Black-box optimization:}
We consider prediction of  users' ratings    in  the  Yahoo! music system \citep{dror2012yahoo}. 
Our  dataset, provided by \citep{lian2016comprehensive},  include $n^\prime = 131072$ true music ratings $\mathbf r \in \mathbb R^{n^\prime}$, and the predicted ratings of   $m = 237$  individual models created 
from the NTU KDD-Cup team \citep{chen2011linear}. 
  Let    $\mathbf C \in \mathbb R^{n \times m}$ represent a matrix of each  models'
predicted ratings on Yahoo! music data sample.
We split the dataset $(\mathbf C, \mathbf r)$ into two equal parts, leading to 
the training dataset  $(\mathbf C_1 \in \mathbb R^{ n\times m }, \mathbf r_1 \in \mathbb R^{n})$ and 
the test dataset $(\mathbf C_2 \in \mathbb R^{ n \times m }, \mathbf r_2 \in \mathbb R^{n})$, where $n = n^\prime/2$.

   \begin{figure}[htb] 
  \centering
 \begin{tabular}{c}
\includegraphics[width=0.4\textwidth,height=!]{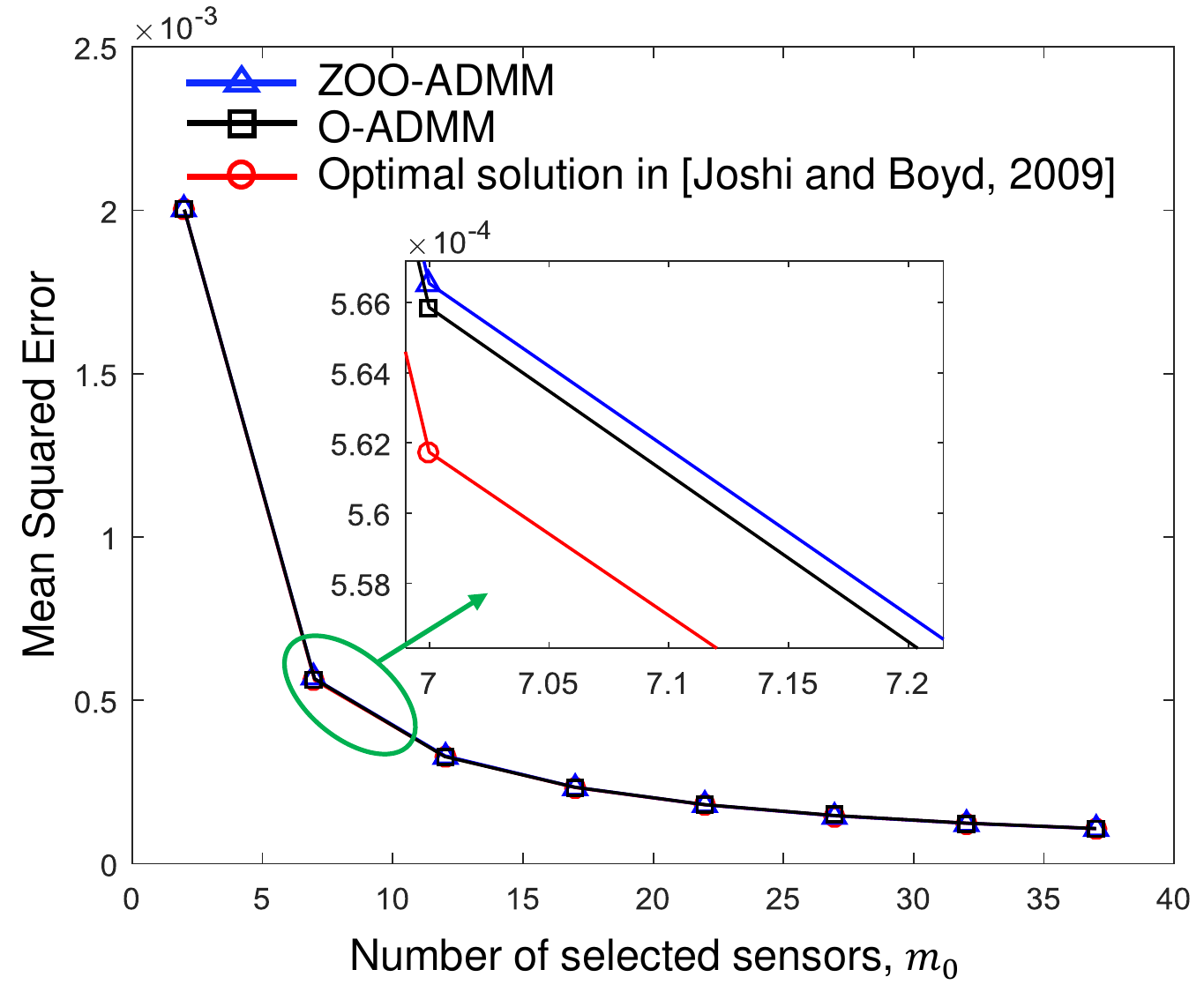} \\
(a) \\
\includegraphics[width=0.415\textwidth,height=!]{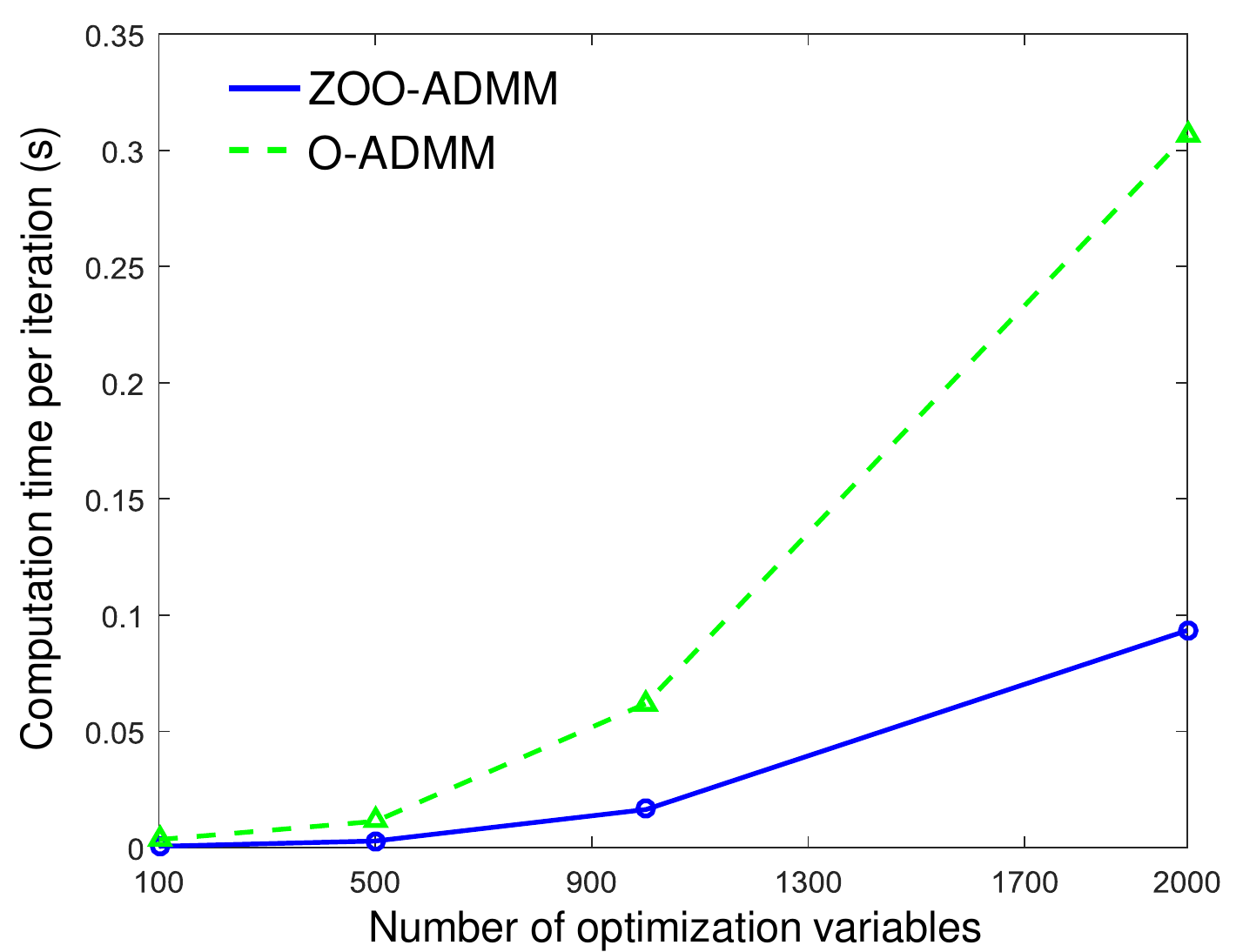}\\
(b)
\end{tabular}
 \caption{ZOO-ADMM for sensor selection: a) MSE versus number of selected sensors $m_0$, b) computation time versus number of optimization variables.}
\label{fig: sensr}
\end{figure}

Our goal is to find the optimal coefficients $\mathbf x $ to blend $m$ individual models such that  the mean squared error $f(\mathbf x) \Def \frac{1}{n} \sum_{i=1}^n f(\mathbf x;\mathbf w_i) =  \frac{1}{n} \sum_{i=1}^n ( [\mathbf C_1]_{i}^T \mathbf x -  [\mathbf r_1]_i )^2$ is minimized, 
where
 $\mathbf w_i = ([\mathbf C_1]_{i}, [\mathbf r_1]_i)$,
 $[\mathbf C_1]_{i}$ is the $i$th row vector of $\mathbf C_1$,  and $[\mathbf r_1]_i $ is the $i$th entry of $\mathbf r_1$.
\textcolor{black}{Since $(\mathbf C, \mathbf r)$ includes  predicted ratings on Yahoo! Music data using NTU KDD-Cup team's models, it is private information known only to other users. Therefore,  the information $(\mathbf C, \mathbf r)$ cannot be accessed directly \citep{lian2016comprehensive}, and explicit gradient calculation for   $f$ is not possible.} We thus treat  the loss function  as a black box, where it is evaluated at individual points $\mathbf x$ in its domain but not over any open region of its domain.
 
  As discussed in Sec.\,\ref{sec: black},  
we can  apply ZOO-ADMM to solve  the proposed linear blending problem, and the prediction accuracy can be measured by   the root mean squared error (RMSE) of the test data
$
\mathrm{RMSE} = \sqrt{ \| \mathbf r_2 - \mathbf C_2 \mathbf x  \|_2^2/n }
$, where an update of $\mathbf x$   is obtained at each iteration.

 In Fig.\,\ref{fig: black}, we compare the   performance of ZO-ADMM with O-ADMM and the optimal solution provided by \citep{lian2016comprehensive}.  
 In Fig.\,\ref{fig: black}-(a), we present RMSE as a function of iteration number 
 under different minibatch schemes.  As we can see, 
both gradient sample averaging  (over $\{ \mathbf z_t \}$) and observation averaging  (over $\{ \mathbf w_t \}$)  significantly accelerate the convergence speed of ZOO-ADMM. In particular, when the minibatch size $q$ is large enough ($50$ in our example),  the dimension-dependent slowdown  factor of ZOO-ADMM  can be mitigated. We also observe that 
 ZOO-ADMM reaches the best   RMSE in \citep{lian2016comprehensive}  after $10000$ iterations. 
 In Fig.\,\ref{fig: black}-(b), we show the convergence error $\| \mathbf x_{t+1} - \mathbf x_t \|_2$ versus iteration number   using  gradient sample averaging minibatch of size $50$. 
  Compared to O-ADMM, ZOO-ADMM   has a larger performance gap in its first few iterations, but it thereafter converges quickly resulting in   comparable performance to O-ADMM.  

 \textbf{Sensor selection:}
We consider an example   of 
estimating a spatial random  field based on measurements of the field at a discrete set of sensor locations. Assume that 
  $m = 100 $ sensors  are randomly deployed over a square region    to   monitor a vector of field intensities    (e.g., temperature values).
    The objective is to estimate the field intensity 
  at  $n = 5$  locations  over a time period of   $T = 1000$ secs.
In \eqref{eq: prob_sensorSel_online}, the observation vectors $\{ \mathbf a_{i,t} \}$ are chosen randomly,
and independently, from a distribution $\mathcal N (\mathbf \mu_i \mathbf 1_n, \mathbf I_n)$. Here $\mu_i$ is generated by an exponential model \citep{liu2016sensor},
$\mu_i = 5 e^{ \sum_{j = 1}^n \| \hat {\mathbf s}_j - \tilde {\mathbf s}_i \|_2/n }
$, where $\hat {\mathbf s}_j $ 
is the $j$-th spatial location at which the field intensity is to be estimated and $\tilde {\mathbf s}_i$ is the spatial location of the $i$ sensor.




In Fig.\,\ref{fig: sensr}, we present the performance of ZOO-ADMM for sensor selection.
In Fig.\,\ref{fig: sensr}-(a), we show
the mean squared error (MSE) averaged over $50$ random trials for different number of selected sensors $m_0$ in \eqref{eq: prob_sensorSel_online}. 
We compare our approach with   O-ADMM and the  method  in \citep{joshi2009sensor}.
The figure shows that ZOO-ADMM yields almost the  same MSE  as O-ADMM. The  method  in \citep{joshi2009sensor} yields slightly better estimation performance, since it uses the second-order optimization method for sensor selection.
 In Fig.\,\ref{fig: sensr}-(b), we present the computation time of ZOO-ADMM versus the number of optimization variables $m$. The figure shows that ZOO-ADMM becomes much more computationally efficient as $m$ increases since no matrix inversion is required.

    \begin{figure} 
  \centering
\includegraphics[width=.4\textwidth]{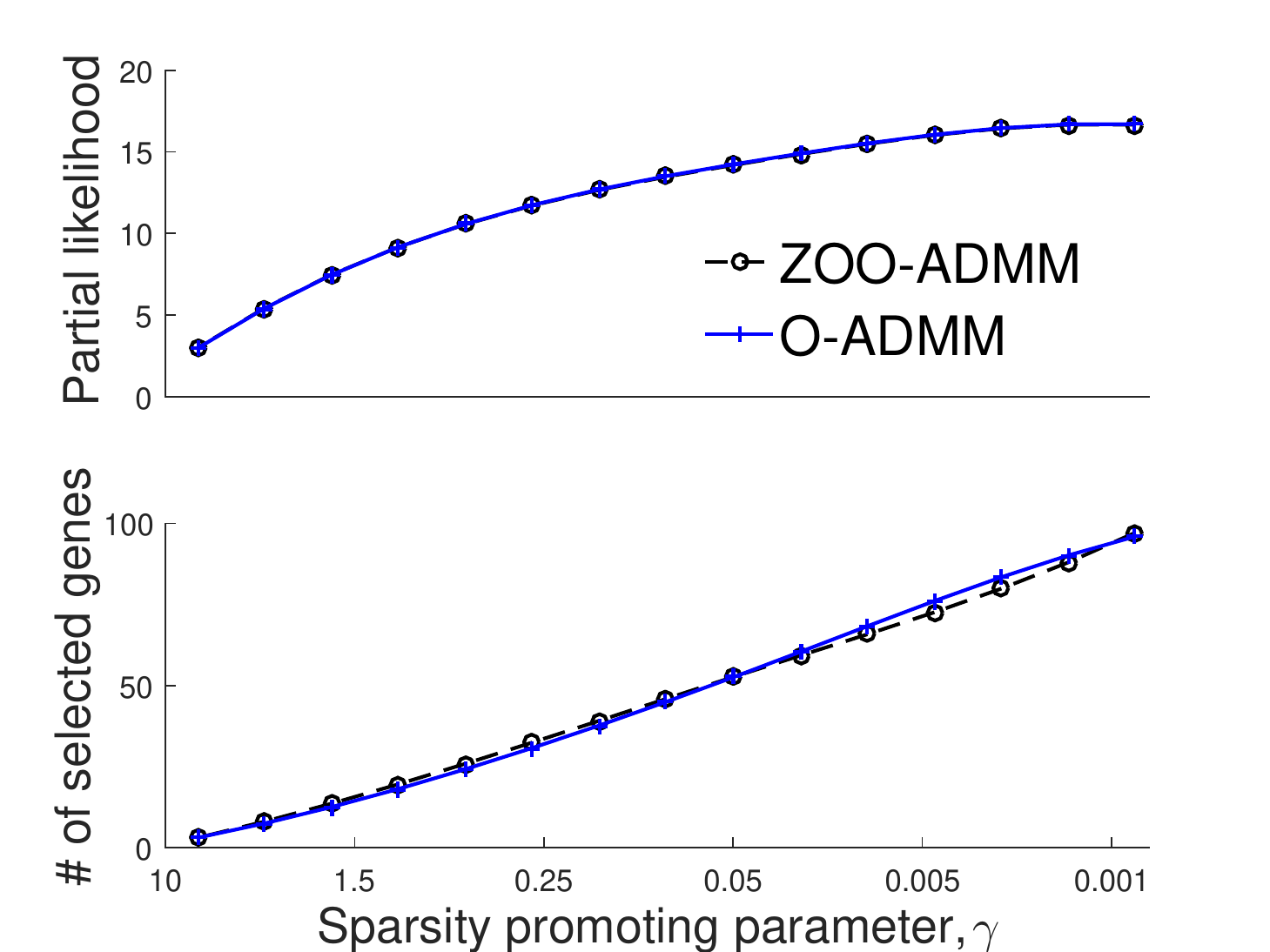}
\caption{{Partial likelihood and number of selected genes versus sparsity promoting parameter $\gamma$.}}
\label{fig: cox}
\end{figure}

\textbf{Sparse Cox regression:}
We next employ ZOO-ADMM to solve problem \eqref{eq: cox_reg_ori} for building a sparse predictor of patient survival using the Kidney renal clear cell carcinoma   dataset\footnote{Available at  \url{http://gdac.broadinstitute.org/}}. The aforementioned dataset includes    clinical data (survival time and censoring information) and  gene expression data for $606$ patients ($534$   with tumor and $72$    without tumor). 
Our goal is to seek the best subset of genes (in terms of optimal sparse covariate coefficients)
that make the most significant impact on  the  survival time.  

\begin{table}[]
\centering
\caption{{Percentage of common genes found using  ZOO-ADMM and Cox scores  \citep{witten2010survival}.}}
\label{my-label}
\begin{tabular}{ |c | c | c | c | }
    \hline
 &   $\gamma = 1.5 $  & $\gamma = 0.05 $   &  $\gamma = 0.001 $  \\ \hline
\# selected genes &   19 & 56 & 93 \\ \hline
Overlapping ($\%$) &   80.1\% & 87.5\% & 92.3\% \\ \hline
    \end{tabular}
\end{table}

   In Fig.\,\ref{fig: cox}, we show the   partial likelihood and number of selected genes as functions of  the  regularization parameter   $\gamma$. 
The figure shows that ZOO-ADMM   
nearly attains the accuracy of  O-ADMM. Furthermore, the   likelihood increases as 
   the number of selected genes  increases. There is thus a   tradeoff   between the (negative) log partial  likelihood and the sparsity of covariate coefficients   in problem \eqref{eq: cox_reg_ori}.   
   To   test the significance of our selected genes, we compare our approach with the  significance analysis  based on univariate Cox scores used in \citep{witten2010survival}. 
The percentage of overlap between the  genes identified by each method  is shown in 
 Table\,\ref{my-label} under different values of $\gamma$. 
 Despite its use of a zeroth order approximation to the gradient, the ZOO-ADMM selects at least $80\%$ of the genes selected by the gradient-based Cox  scores of \citep{witten2010survival}.
 

\section{Conclusion}
\label{sec: conc}
In this paper, we proposed  and analyzed a gradient-free (zeroth-order)  online  optimization  algorithm, ZOO-ADMM.
  We showed that the regret bound of ZOO-ADMM  
  suffers an additional dimension-dependent factor   in convergence
rate over gradient-based  online variants of ADMM, leading to $O(\sqrt{m}/\sqrt{T})$ convergence rate, where $m$ is the number of optimization variables. 
To alleviate the dimension dependence, we presented 
two minibatch strategies that yield an improved convergence rate of  $O(\sqrt{1+q^{-1}m}/\sqrt{T})$, where $q$ is the minibatch size.  We   illustrated the effectiveness of ZOO-ADMM via multiple applications  using both synthetic   and real-world datasets.  
In the future,
we would like to relax the assumptions on   smoothness and    convexity   of the    cost function in   ZOO-ADMM. 

\subsubsection*{Acknowledgements}

This work was partially supported by grants from the US Army Research
Office, grant numbers W911NF-15-1-0479 and W911NF-15-1-0241. The work of J. Chen was supported in part by the Natural Science Foundation of China under Grant 61671382.

\clearpage

\section{Supplementary Material}
\subsection{Assumptions and Key Notations}
\label{sec: assump}

Recall that we  consider the   regularized loss minimization   problem  over a time horizon of length $T$,
\begin{align}
\begin{array}{ll}
    \displaystyle \minimize_{\mathbf x \in \mathcal X, \mathbf y \in \mathcal Y} &  \displaystyle  \frac{1}{T} \sum_{t=1}^T  f_t(\mathbf x; \mathbf w_t) 
    + \phi(\mathbf y) \\
   \st  & \mathbf A \mathbf x + \mathbf B \mathbf y = \mathbf c.
\end{array}
\label{eq: prob_online_reg_xy}
\end{align}
ZOO-ADMM is given by
\begin{align}
& \mathbf x_{t+1} = \argmin_{\mathbf x \in \mathcal X} \left \{  \hat {\mathbf g}_t^T \mathbf x - \boldsymbol \lambda_t^T (\mathbf A \mathbf x + \mathbf B \mathbf y_t - \mathbf c) \right. \nonumber \\
& \left. + \frac{\rho}{2} \left \| \mathbf A \mathbf x + \mathbf B \mathbf y_t - \mathbf c \right \|_2^2 + \frac{1}{2 \eta_t}  \| \mathbf x - \mathbf x_t \|_{\mathbf G_t}^2 \right \}, \label{eq: x_step_ZOADMM}   \\
& \mathbf y_{t+1} = \argmin_{\mathbf y \in \mathcal Y} \left  \{  \phi(\mathbf y) -  \boldsymbol \lambda_t^T (\mathbf A \mathbf x_{t+1} + \mathbf B \mathbf y  - \mathbf c)  \right. \nonumber \\
& \left. + \frac{\rho}{2} \| \mathbf A \mathbf x_{t+1} + \mathbf B \mathbf y  - \mathbf c\|_2^2  \right \}, \label{eq: y_step} \\
& \boldsymbol \lambda_{t+1} =  \boldsymbol \lambda_t - \rho (\mathbf A \mathbf x_{t+1} + \mathbf B \mathbf y_{t+1}  - \mathbf c), \label{eq: dual_step}
\end{align}
where $\mathbf G_t = \alpha \mathbf I - \rho \eta_t \mathbf A^T \mathbf A$.

We first elaborate on our assumptions. 
\begin{itemize}
\item Assumption A implies that   $\| \mathbf x - \mathbf x^{\prime} \|_2 \leq R$ and $\| \mathbf y - \mathbf y^{\prime} \|_2 \leq R$ for all $\mathbf x, \mathbf x^{\prime} \in \mathcal X$ and for all $\mathbf y, \mathbf y^{\prime} \in \mathcal Y$.  
\item 
Based on Jensen's inequality, Assumptions B  implies that
$\|\mathbb E [ \nabla_{\mathbf x} f(\mathbf x;\mathbf w_t)] \|_2 \leq L_1$.
\item 
Assumption C implies a Lipschitz condition over the gradient $\nabla_{\mathbf x} f(\mathbf x; \mathbf w_t)$ with constant $L_{g} (\mathbf w_t)$ \citep{bubeck2015convex,hazan2016introduction}. Also based on Jensen's inequality, we have
$
| \mathbb E[ L_{g}(\mathbf w_t)  ] | \leq L_g
$.
\end{itemize}

We next introduce key notations used in our analysis. 
Given the primal-dual variables $\mathbf x$, $\mathbf y$ and $\boldsymbol \lambda$ of problem \eqref{eq: prob_online_reg_xy}, we define 
${\mathbf v} \Def [ \mathbf x^T, \mathbf y^T,  {\boldsymbol \lambda}^T]$, 
 and a primal-dual mapping $H$
\begin{align}
H(\mathbf v) \Def 
 \mathbf C \mathbf v - \begin{bmatrix}
0 \\
0 \\
\mathbf  c 
\end{bmatrix}, ~ \mathbf C \Def \begin{bmatrix}
0 & 0 & - \mathbf A^T \\
0 & 0 & - \mathbf B^T \\
\mathbf A & \mathbf B & 0
\end{bmatrix}, \label{eq: PD_map}
\end{align}
where  $\mathbf C$ is  skew symmetric, namely, $\mathbf C^T = - \mathbf C$. An important property of the affine mapping $H$ is that 
$
\langle \mathbf v_1 - \mathbf v_2, H(\mathbf v_1) -  H(\mathbf v_2) \rangle = 0 
$ for every $\mathbf v_1$ and $\mathbf v_2$.
Supposing the sequence $\{ \mathbf v_t \}$ is generated by an algorithm, we introduce the
auxiliary sequence 
\begin{align}
\tilde{\mathbf v}_t \Def [ \mathbf x_t^T, \mathbf y_t^T,  \tilde{\boldsymbol \lambda}_t^T]^T, \label{eq: vtilde}
\end{align}
where $\tilde{\boldsymbol \lambda}_t \Def {\boldsymbol \lambda}_t - \rho (\mathbf A \mathbf x_{t+1} + \mathbf B \mathbf y_t - \mathbf c) $.

\subsection{Proof of Theorem\,1}
\label{sec: thr1}
Since  the sequences $\{ \mathbf x_t \}$, $\{ \mathbf y_t \}$ and $\{ \boldsymbol \lambda_t \}$ produced from \eqref{eq: x_step_ZOADMM}-\eqref{eq: dual_step} have the same structure as the ADMM/O-ADMM steps,   the property of ADMM given by Theorem\,4 of \citep{suzuki2013dual} is directly applicable to our case, yielding
\begin{align}
&  \sum_{t=1}^T ( f_t(\mathbf x_t) + \phi(\mathbf y_t) )-  \sum_{t=1}^T ( f_t(\mathbf x) + \phi(\mathbf y) ) \nonumber \\
&  + \sum_{t=1}^T (\tilde{\mathbf v}_t - \mathbf v)^T H(\tilde{\mathbf v}_t )   
 \leq  \frac{\| \mathbf x_1  - \mathbf x\|_{\mathbf G_1}^2}{2\eta_1}  \nonumber \\
& + \sum_{t = 2}^T \left ( \frac{\| \mathbf x_t - \mathbf x \|_{\mathbf G_t}^2}{2\eta_t} - \frac{\| \mathbf x_t - \mathbf x \|_{\mathbf G_{t-1}}^2}{2\eta_{t-1}} \right )  \nonumber \\
& + \inp{\boldsymbol \lambda}{\mathbf A (\mathbf x_{T+1} - \mathbf x_1)} + \frac{\rho}{2} \| \mathbf y_1 - \mathbf y \|_{\mathbf B^T \mathbf B} + \frac{\|  \boldsymbol \lambda_1 - \boldsymbol \lambda \|_2^2}{2\rho} \nonumber \\
&  - \frac{\|  \boldsymbol \lambda_{T+1} - \boldsymbol \lambda \|_2^2}{2\rho} + \inp{\mathbf B (\mathbf y - \mathbf y_{T+1})}{\boldsymbol \lambda_{T+1} - \boldsymbol \lambda} \nonumber \\ 
 & - \inp{\mathbf B (\mathbf y - \mathbf y_1)}{\boldsymbol \lambda_1 -\boldsymbol \lambda } - \sum_{t=1}^T \frac{\| \lambda_t - \lambda_{t+1}\|_2^2}{2\rho} \nonumber \\
& - \sum_{t=1}^T \frac{\sigma}{2} \| \mathbf x_t - \mathbf x \|_2^2 + \sum_{t=1}^T \frac{\eta_t}{2} \| \hat {\mathbf g}_t \|_{\mathbf G_t^{-1}}^2. \label{eq: property_OADMM}
\end{align}
Here     for notational simplicity we have used, and henceforth
will continue to use, $f_t(\mathbf x_t)$ instead of $f(\mathbf x_t; \mathbf w_t)$.

In \eqref{eq: property_OADMM},  based on $ \mathbf G_t = \alpha \mathbf I - \rho \eta_t \mathbf A^T \mathbf A$, we have
\begin{align}
& \frac{\| \mathbf x_t - \mathbf x \|_{\mathbf G_t}^2}{2\eta_t} - \frac{\| \mathbf x_t - \mathbf x \|_{\mathbf G_{t-1}}^2}{2\eta_{t-1}} \nonumber \\
= & \left ( 
 \frac{\alpha}{2\eta_t} - \frac{\alpha}{2\eta_{t-1}} 
 \right )\| \mathbf x_t - \mathbf x \|_2^2 , \nonumber 
\end{align}
which yields
\begin{align}
& \sum_{t = 2}^T \left ( \frac{\| \mathbf x_t - \mathbf x \|_{\mathbf G_t}^2}{2\eta_t} - \frac{\| \mathbf x_t - \mathbf x \|_{\mathbf G_{t-1}}^2}{2\eta_{t-1}} \right )  \nonumber \\
&-  \sum_{t=1}^T \frac{\sigma}{2} \| \mathbf x_t - \mathbf x \|_2^2 \leq \nonumber \\
& \sum_{t=2}^T \max\{  \frac{\alpha}{2\eta_t} - \frac{\alpha}{2\eta_{t-1}}  - \frac{\sigma}{2} , 0 \} R^2. \label{eq: appA_eq1}
\end{align}

We also note that the terms 
$
\frac{1}{2\eta_1}\| \mathbf x_1  - \mathbf x\|_{\mathbf G_1}^2
$, $\inp{\boldsymbol \lambda}{\mathbf A (\mathbf x_{T+1} - \mathbf x_1)}$,
$\frac{\rho}{2} \| \mathbf y_1 - \mathbf y \|_{\mathbf B^T \mathbf B}
$,
$
\frac{1}{2\rho} ( \|  \boldsymbol \lambda_1 - \boldsymbol \lambda \|_2^2 -  \|  \boldsymbol \lambda_{T+1} - \boldsymbol \lambda \|_2^2 )
$,
$
\inp{\mathbf B (\mathbf y - \mathbf y_{T+1})}{\boldsymbol \lambda_{T+1} - \boldsymbol \lambda}
$, and
$
\inp{\mathbf B (\mathbf y - \mathbf y_1)}{\boldsymbol \lambda_1 -\boldsymbol \lambda }
$ are \textit{independent} of time $t$.
In particular, we have 
\begin{align}
&\| \mathbf x_1  - \mathbf x \|_{\mathbf G_1}^2  \leq \alpha R^2, \nonumber \\
& \inp{\boldsymbol \lambda}{\mathbf A (\mathbf x_{T+1} - \mathbf x_1)} 
\leq R \| \boldsymbol \lambda \|_2  \| \mathbf A \|_F , \nonumber \\
& ( \|  \boldsymbol \lambda_1 - \boldsymbol \lambda \|_2^2 -  \|  \boldsymbol \lambda_{T+1} - \boldsymbol \lambda \|_2^2 ) \leq  \| \boldsymbol \lambda \|_2^2, \nonumber \\
& \inp{\mathbf B (\mathbf y - \mathbf y_1)}{\boldsymbol \lambda -\boldsymbol \lambda_1 } \leq R \| \mathbf B \|_F \| \boldsymbol \lambda \|_2,\label{eq: appA_eq2}
\end{align}
where $\| \cdot \|_F$ denotes the Frobenius norm of a matrix, and we have used the facts that $\mathbf G_t \preceq \alpha \mathbf I$ and $\boldsymbol \lambda_1 = \mathbf 0$.

Based on the optimality condition of  $\mathbf y_{t+1}$ in \eqref{eq: y_step}, we have
$
\inp{  \partial \phi(\mathbf y_{t+1} ) - \mathbf B^T \boldsymbol \lambda_t + \rho \mathbf B^T ( \mathbf A \mathbf x_{t+1} + \mathbf B \mathbf y_{t+1} - \mathbf c ) }{ \mathbf y - \mathbf y_{t+1} } \geq 0~, \forall \mathbf y \in \mathcal Y$,
which is equivalent to $\inp{  \partial \phi(\mathbf y_{t+1} ) - \mathbf B^T \boldsymbol \lambda_{t+1}  }{ \mathbf y - \mathbf y_{t+1} } \geq 0 $. And thus, we obtain
\begin{align}
 & \inp{   \boldsymbol \lambda_{t+1}  }{ \mathbf B( \mathbf y - \mathbf y_{t+1} ) } - \inp{   \boldsymbol \lambda  }{ \mathbf B( \mathbf y - \mathbf y_{t+1} ) }  \nonumber \\
&  \leq \inp{  \partial \phi(\mathbf y_{t+1} ) }{ \mathbf y - \mathbf y_{t+1} } - \inp{   \boldsymbol \lambda  }{ \mathbf B( \mathbf y - \mathbf y_{t+1} ) } ,\nonumber  \end{align}
 which yields
 \begin{align}
 & \inp{\mathbf B (\mathbf y - \mathbf y_{t+1})}{\boldsymbol \lambda_{t+1} - \boldsymbol \lambda}  \nonumber \\
 \leq & \inp{\mathbf y - \mathbf y_{t+1}}{ \partial \phi(\mathbf y_{t+1}) - \mathbf B^T \boldsymbol \lambda } \nonumber \\
\leq  & R( L_2 + \| \mathbf B^T \boldsymbol \lambda \|_2 ),  \label{eq: appA_eq3}
\end{align} 
where we have used the fact that $\| \partial \phi(\mathbf y_{t+1})  \|_2 \leq L_2$.

Substituting \eqref{eq: appA_eq1}-\eqref{eq: appA_eq3} into \eqref{eq: property_OADMM}, we then obtain
\begin{align}
&  \frac{1}{T}  \sum_{t=1}^T \left ( f_t(\mathbf x_t) + \phi(\mathbf y_t) \right )-  \frac{1}{T}  \sum_{t=1}^T \left ( f_t(\mathbf x) + \phi(\mathbf y) \right ) \nonumber \\
&  + \frac{1}{T}  \sum_{t=1}^T (\tilde{\mathbf v}_t - \mathbf v)^T H(\tilde{\mathbf v}_t )    +  \frac{1}{T} \sum_{t=1}^T \frac{\|  \boldsymbol \lambda_{t+1} - \boldsymbol \lambda_{t}\|_2^2}{2\rho} \nonumber \\
& \leq  \frac{1}{T} \sum_{t=2}^T \max\{  \frac{\alpha}{2\eta_t} - \frac{\alpha}{2\eta_{t-1}}  - \frac{\sigma}{2} , 0 \} R^2  \nonumber \\
& + \frac{1}{T}\sum_{t=1}^T \frac{\eta_t}{2} \|  \hat {\mathbf g}_t  \|^2 + \frac{K}{T},
\label{eq: prop_ZOADMM}
\end{align}
where 
$K$ is a constant term related to $\alpha$, $R$, $\eta_1$, $\mathbf A$, $\mathbf B$, $\boldsymbol \lambda$, $\rho$ and $L_2$,
$
K = \frac{\alpha R^2}{2\eta_1 } + R \| \boldsymbol \lambda \|_2  \| \mathbf A \|_F + \frac{1}{2\rho} \| \boldsymbol \lambda \|_2^2 + R \| \mathbf B \|_F \| \boldsymbol \lambda \|_2 + R( L_2 + \| \mathbf B^T \boldsymbol \lambda \|_2 )
$, and we have used the fact that  $\|  \hat {\mathbf g}_t  \|_{\mathbf G_t^{-1}}^2 \leq \|  \hat {\mathbf g}_t  \|_{2}^2$ (due to $\mathbf G_t^{-1} \preceq \mathbf I$).

Based on \eqref{eq: prop_ZOADMM}
we continue to prove Theorem\,1.
When $\mathbf B$ is invertible and $\mathbf y_t^\prime = \mathbf B^{-1} (\mathbf c - \mathbf A \mathbf x_t)$, we obtain 
\begin{align} \label{eq: appc_1}
\mathbf B ({\mathbf y}_t^\prime  - {\mathbf y}_t ) = \frac{1}{\rho} ( \boldsymbol \lambda_t - \boldsymbol \lambda_{t-1} ).
\end{align}
Based on the convexity of  $f$ and $\phi$, we obtain
\begin{align}  \label{eq: appc_2}
& f_t({\mathbf x}_t) + \phi({\mathbf y}_t^\prime)
 \leq f_t({\mathbf x}_t)  + \phi({\mathbf y}_t) + \inp{\partial \phi({\mathbf y}_t^\prime)}{{\mathbf y}_t^\prime - {\mathbf y}_t} \nonumber \\
& = f_t({\mathbf x}_t)  + \phi({\mathbf y}_t) +\frac{1}{\rho} \inp{(\mathbf B^{-1} )^T \partial \phi({\mathbf y}_t^\prime)}{ \boldsymbol \lambda_t - \boldsymbol \lambda_{t-1}},
\end{align}
where 
the last equality holds due to \eqref{eq: appc_1}.

Let $(\mathbf x^*, \mathbf y^*)$  be the optimal solution (implying $\mathbf A \mathbf x^* + \mathbf B \mathbf y^* - \mathbf c = \mathbf 0$).
For any dual variable $\boldsymbol \lambda^*$ and   $\tilde {\mathbf v}_t 
= [ \mathbf x_t^T, \mathbf y_t^T,  \tilde{\boldsymbol \lambda}_t^T]^T
$, we have
\begin{align}
&(\tilde{\mathbf v}_t - \mathbf v^*)^T H(\tilde{\mathbf v}_t ) =
H(\mathbf v^* )^T(\tilde{\mathbf v}_t - \mathbf v^*) \nonumber \\
= &
\begin{bmatrix}
 - \mathbf A^T \boldsymbol \lambda^* \\
 - \mathbf B^T \boldsymbol \lambda^* \\
 \mathbf A \mathbf x^* +  \mathbf B \mathbf y^* - \mathbf c
\end{bmatrix}^T
\begin{bmatrix}
 \mathbf x_t - \mathbf x^* \\
 \mathbf y_t - \mathbf y^* \\
 \tilde{\boldsymbol \lambda}_t - \boldsymbol \lambda^*
\end{bmatrix} \nonumber \\\
= & \inp{\boldsymbol \lambda^*}{\mathbf c - \mathbf A \mathbf x_t -  \mathbf B \mathbf y_t } = \frac{1}{\rho}\inp{\boldsymbol \lambda^*}{\boldsymbol \lambda_t - \boldsymbol \lambda_{t-1} }
\label{eq: appc_3}
\end{align}
where $\mathbf v^*  \Def [ (\mathbf x^*)^T, (\mathbf y^*)^T, (\boldsymbol \lambda^*)^T  ]^T$, and the affine mapping $H(\cdot)$ is given by \eqref{eq: PD_map}. 

Setting $\boldsymbol \lambda^* =  (\mathbf B^{-1} )^T \partial \phi({\mathbf y}_t^\prime)$, based on \eqref{eq: appc_2} and \eqref{eq: appc_3}  we have
\begin{align} \label{eq: appc_4}
&
 f_t({ \mathbf x}_t) + \phi({\mathbf y}_t^\prime) 
-  \left ( f_t( \mathbf x^*) + \phi(\mathbf y^*) \right )
\nonumber \\
\leq & 
f_t({\mathbf x}_t )  + \phi({\mathbf y}_t) + (\tilde{\mathbf v}_t - \mathbf v^*)^T H(\tilde{\mathbf v}_t ) \nonumber \\
& - ( f_t( \mathbf x^* ) + \phi(\mathbf y^*) ).
\end{align}
Combining  \eqref{eq: prop_ZOADMM} and \eqref{eq: appc_4} yields
\begin{align}
&  \frac{1}{T}  \sum_{t=1}^T \left ( f_t(\mathbf x_t) + \phi(\mathbf y_t^\prime) \right )-  \frac{1}{T}  \sum_{t=1}^T \left ( f_t(\mathbf x^*) + \phi(\mathbf y^*) \right )   \nonumber \\
&  +  \frac{1}{T} \sum_{t=1}^T \frac{\|  \boldsymbol \lambda_{t+1} - \boldsymbol \lambda_{t}\|_2^2}{2\rho}  
\leq \frac{1}{T}  \sum_{t=1}^T \left ( f_t({\mathbf x}_t)  + \phi({\mathbf y}_t) \right)  \nonumber \\
& - \frac{1}{T}  \sum_{t=1}^T \left ( f_t( \mathbf x^*) + \phi(\mathbf y^*) \right )  + \frac{1}{T} \sum_{t=1}^T  (\tilde{\mathbf v}_t - \mathbf v^*)^T H(\tilde{\mathbf v}_t )  \nonumber \\
& +  \frac{1}{T} \sum_{t=1}^T \frac{\|  \boldsymbol \lambda_{t+1} - \boldsymbol \lambda_{t}\|_2^2}{2\rho} \nonumber \\
\leq &
\frac{1}{T} \sum_{t=2}^T \max\{  \frac{\alpha}{2\eta_t} - \frac{\alpha}{2\eta_{t-1}}  - \frac{\sigma}{2} , 0 \} R^2 \nonumber \\
& + \frac{1}{T}\sum_{t=1}^T \frac{\eta_t}{2} \|  \hat {\mathbf g}_t  \|_2^2 + \frac{K}{T}.
\label{eq: appc_5}
\end{align}
Since $\boldsymbol \lambda_{t+1} - \boldsymbol \lambda_{t} = \rho (\mathbf A \mathbf x_{t+1} + \mathbf B \mathbf y_{t+1} - \mathbf c)$, from \eqref{eq: appc_5} we have
\begin{align}\label{eq: appc_6}
 &  \frac{1}{T}  \sum_{t=1}^T \left ( f_t(\mathbf x_t) + \phi(\mathbf y_t^\prime) \right )-  \frac{1}{T}  \sum_{t=1}^T \left ( f_t(\mathbf x^*) + \phi(\mathbf y^*) \right )  \nonumber \\
&  +  \frac{\rho}{2T} \sum_{t=1}^T \|\mathbf A \mathbf x_{t+1} + \mathbf B \mathbf y_{t+1} - \mathbf c\|_2^2
\nonumber \\
\leq & \frac{1}{T} \sum_{t=2}^T \max\{  \frac{\alpha}{2\eta_t} - \frac{\alpha}{2\eta_{t-1}}  - \frac{\sigma}{2} , 0 \} R^2 \nonumber \\
& + \frac{1}{T}\sum_{t=1}^T \frac{\eta_t}{2} \|  \hat {\mathbf g}_t  \|_2^2 + \frac{K}{T}.
\end{align}

Taking expectations for both sides of \eqref{eq: appc_6} with respect to its randomness, we have 
\begin{align}\label{eq: appc_7}
&\mathbb E \left [
\frac{1}{T}  \sum_{t=1}^T \left ( f_t(\mathbf x_t) + \phi(\mathbf y_t^\prime) \right )-  \frac{1}{T}  \sum_{t=1}^T \left ( f_t(\mathbf x^*) + \phi(\mathbf y^*) \right )  \right ]  \nonumber \\
&  + \mathbb E \left [  \frac{\rho}{2T} \sum_{t=1}^T \|\mathbf A \mathbf x_{t+1} + \mathbf B \mathbf y_{t+1} - \mathbf c\|_2^2 \right ] \nonumber \\
\leq &
\frac{1}{T} \sum_{t=2}^T \max\{  \frac{\alpha}{2\eta_t} - \frac{\alpha}{2\eta_{t-1}}  - \frac{\sigma}{2} , 0 \} R^2  \nonumber \\
& +   \frac{1}{T}\sum_{t=1}^T \frac{\eta_t}{2} \mathbb E [ \|  \hat {\mathbf g}_t  \|_2^2 ] + \frac{K}{T}.
\end{align}

Based on \citep[Lemma\,1]{duchi2015optimal}, the second-order statistics of the gradient estimate $\hat{\mathbf g}_t $ is given by
\begin{align}
&\mathbb E_{\mathbf z_t}[\hat{\mathbf g}_t] = \mathbf g_t  
+ \beta_t L_{g}(\mathbf w_t) \nu(\mathbf x_t, \beta_t),
\label{eq: mu_gradient} \\
&\mathbb E_{\mathbf z_t}[\| \hat{\mathbf g}_t \|_2^2] \leq 2 s(m) \| \mathbf g_t\|_2^2 + \frac{1}{2} \beta_t^2 L_{g}(\mathbf w_t)^2 M(\mu)^2, \label{eq: var_gradient}
\end{align}
where $\mathbf g_t = \nabla_{\mathbf x} f(\mathbf x;\mathbf w_t) |_{\mathbf x = \mathbf x_t}$, $\| \nu(\mathbf x_t, \beta_t)  \|_2 \leq \frac{1}{2} \mathbb E_{\mathbf z}[ \| \mathbf z \|_2^3  ]$,  $L_{g}(\mathbf w_t)$ is defined in Assumption C, and $s(m)$ and $M(\mu)$ are introduced in Assumption E.
According to \eqref{eq: var_gradient}, we have
\begin{align}\label{eq: appc_8}
&  \mathbb E [ \|  \hat {\mathbf g}_t  \|_2^2 ] = \mathbb E \left [
 \mathbb E_{\mathbf z} [\|  \hat {\mathbf g}_t  \|_2^2 ]
 \right ] \nonumber \\
\leq  & \mathbb E \left [ 
 2 s(m) \| \mathbf g_t\|_2^2 + \frac{1}{2} \beta_t^2 L_{g,t}^2 M(\mu)^2
 \right ]
\nonumber  \\
\leq & 2 s(m)  L_1^2+ \frac{1}{2} \beta_t^2 L_g^2 M(\mu)^2, 
\end{align}
where for ease of notation, we have replaced $L_g(\mathbf w_t)$ with $L_{g,t}$, and
the last inequality holds due to Assumptions B and C. 

Substituting \eqref{eq: appc_8} into \eqref{eq: appc_7},
 the expected average regret can be bounded as
\begin{align}\label{eq: appc_10}
& \overline{\mathrm{Regret}}_{T}(  \mathbf x_t, \mathbf y_t^\prime, \mathbf x^*, \mathbf y^* ) \nonumber \\
  \leq  &
\frac{1}{T} \sum_{t=2}^T \max\{  \frac{\alpha}{2\eta_t} - \frac{\alpha}{2\eta_{t-1}}  - \frac{\sigma}{2} , 0 \} R^2 +     \frac{s(m) L_1^2}{T}\sum_{t=1}^T \eta_t  \nonumber \\
& + \frac{M(\mu)^2 L_g^2}{4T}\sum_{t=1}^T \eta_t \beta_t^2 + \frac{K}{T}.
\end{align}

On the other hand, when $\mathbf A$ is invertible and $\mathbf x_t^\prime = \mathbf A^{-1} (\mathbf c - \mathbf B \mathbf y_t)$, we obtain
\begin{align} 
\mathbf A ({\mathbf x}_t^\prime  - {\mathbf x}_t ) = \frac{1}{\rho} ( \boldsymbol \lambda_t - \boldsymbol \lambda_{t-1} ). \nonumber
\end{align}
Based on the convexity of  $f$ and $\phi$, we obtain
\begin{align} 
& f_t({\mathbf x}_t^\prime ) + \phi({\mathbf y}_t) \nonumber \\
 \leq & f_t({\mathbf x}_t)  + \phi({\mathbf y}_t) + \inp{\nabla f_t({\mathbf x}_t^\prime)}{{\mathbf x}_t^\prime - {\mathbf x}_t } \nonumber \\
 = & f_t({\mathbf x}_t )  + \phi({\mathbf y}_t) +\frac{1}{\rho} \inp{(\mathbf A^{-1} )^T \nabla f_t({\mathbf x}_t^\prime)}{ \boldsymbol \lambda_t - \boldsymbol \lambda_{t-1}}.
\label{eq: app41}
\end{align}
Setting   $\boldsymbol \lambda^* =  (\mathbf A^{-1} )^T \nabla f_t({\mathbf x}_t^\prime)$, based on \eqref{eq: app41}  and \eqref{eq: appc_3}  we have
\begin{align} \label{eq: app4_2}
&
 f_t({ \mathbf x}_t^\prime) + \phi({\mathbf y}_t) 
-  \left ( f_t( \mathbf x^* ) + \phi(\mathbf y^*) \right ) \leq f_t({\mathbf x}_t) 
\nonumber \\
& 
 + \phi({\mathbf y}_t) + (\tilde{\mathbf v}_t - \mathbf v^*)^T H(\tilde{\mathbf v}_t ) - ( f_t( \mathbf x^* ) + \phi(\mathbf y^*) ).
\end{align}
Since the right hand side (RHS) of \eqref{eq: app4_2} and RHS of \eqref{eq: appc_4} are  same, we can then mimic  the aforementioned procedure   to prove that the regret $\overline{\mathrm{Regret}}_{T}(  \mathbf x_t^\prime, \mathbf y_t, \mathbf x^*, \mathbf y^* )$ obeys the same bounds as 
\eqref{eq: appc_10}.

\subsection{Simplification of Regret Bound}
\label{sec: simplify}
Consider terms in right hand side  (RHS) of \eqref{eq: appc_10} together with
$\eta_t = \frac{C_1}{ \sqrt{s(m)}  \sqrt{t}}$ and 
$\beta_t = \frac{C_2}{{M(\mu)}t}$, 
we have
\begin{align}\label{eq: appc_12}
&\frac{1}{T} \sum_{t=2}^T \max\{  \frac{\alpha}{2\eta_t} - \frac{\alpha}{2\eta_{t-1}}  - \frac{\sigma}{2} ,0 \} R^2  \nonumber \\
 \leq & \frac{1}{T} \sum_{t=2}^T (  \frac{\alpha}{2\eta_t} - \frac{\alpha}{2\eta_{t-1}}  ) R^2  \leq \frac{1}{\sqrt{T}}\frac{\alpha R^2 \sqrt{s(m)}}{2C_1}, \nonumber \\
& \frac{s(m) L_1^2}{T}\sum_{t=1}^T \eta_t   \leq \frac{2 C_1 \sqrt{s(m)} L_1^2}{\sqrt{T}}, \nonumber \\  & \frac{M(\mu)^2 L_g^2}{4T}\sum_{t=1}^T \eta_t \beta_t^2
= \frac{C_1 C_2^2 L_g^2 }{4 \sqrt{s(m)} T } \sum_{t=1}^T \frac{1}{t^{5/2}} \nonumber \\
 & \leq  
\frac{5C_1 C_2^2 L_g^2 }{12T}, 
\end{align} 
 where we have used the facts that $\sum_{t=1}^T \frac{1}{\sqrt{t}} \leq 2 \sqrt{T}$,
\begin{align} \label{eq: appc_13}
 &  \sum_{t=1}^T (1/t^a)  =  1+ \sum_{t=2}^T(1/t^a)   \nonumber \\
 \leq  & 1 + \int_1^{\infty} (1/t^a) = a/(a-1), ~ \forall a > 1,  
\end{align}
and we recall that ${s(m)} = m   \geq 1$. 
Substituting \eqref{eq: appc_12} into RHS of \eqref{eq: appc_10}, we conclude that  the expected average regret $\overline{\mathrm{Regret}}_{T}(  \mathbf x_t, \mathbf y_t^\prime, \mathbf x^*, \mathbf y^* )$ is upper bounded by
\begin{align} \label{eq: regret_ZOADMM_general}
 \frac{1}{\sqrt{T}}\frac{\alpha R^2 \sqrt{s(m)}}{2C_1}  +     \frac{2 C_1 \sqrt{s(m)} L_1^2}{\sqrt{T}}  + \frac{5C_1 C_2^2 L_g^2}{12T} + \frac{K}{T}.  
\end{align}

\subsection{Proof of Corollary\,1}
\label{sec: coro1}

Given i.i.d. samples $\{ \mathbf w_t \}$ 
  drawn from the probability distribution $P$, from Theorem\,1 we have 
\begin{align}\label{eq: appd_1}
&\mathbb E \left [
\frac{1}{T}  \sum_{t=1}^T \left ( f(\mathbf x_t; \mathbf w_t) + \phi(\mathbf y_t^\prime) \right ) \right. \nonumber \\
& \left. -  \frac{1}{T}  \sum_{t=1}^T \left ( f(\mathbf x^*; \mathbf w_t) + \phi(\mathbf y^*) \right )  \right ] \nonumber \\
 \leq &
 \frac{1}{\sqrt{T}}\frac{\alpha R^2 \sqrt{s(m)}}{2C_1}  +     \frac{2 C_1 \sqrt{s(m)} L_1^2}{\sqrt{T}} \nonumber \\
&+ \frac{5C_1 C_2^2 L_g^2}{12} \frac{1}{T}
 + \frac{K}{T}.
\end{align}
Based on $F(\mathbf x, \mathbf y) = \mathbb E_{\mathbf w} [f(\mathbf x; \mathbf w)] + \phi(\mathbf y)$, from \eqref{eq: appd_1} we have
\begin{align}\label{eq: appc_11}
 &\mathbb E \left [  F(\bar{\mathbf x}_t, \bar{\mathbf y}_t) -  F(\mathbf x^*, \mathbf y^*) \right ] \nonumber \\
 \leq  &
 \mathbb E \left [ \frac{1}{T} \sum_{t=1}^T F(\mathbf x_t, \mathbf y_t) -  F(\mathbf x^*, \mathbf y^*) \right ] \nonumber  \\
= &\mathbb E_{\mathbf z_{1:T}} \left [ \mathbb E_{\mathbf w_{1:T}} \left[  \frac{1}{T}  \sum_{t=1}^T \left ( f(\mathbf x_t; \mathbf w_t) + \phi(\mathbf y_t^\prime) \right ) \right. \right. \nonumber \\
& \left. \left.  -  \frac{1}{T}  \sum_{t=1}^T \left ( f(\mathbf x^*; \mathbf w_t) + \phi(\mathbf y^*) \right )  \right ] \right ] \nonumber  \\
\leq & \frac{1}{\sqrt{T}}\frac{\alpha R^2 \sqrt{s(m)}}{2C_1}  +     \frac{2 C_1 \sqrt{s(m)} L_1^2}{\sqrt{T}}  \nonumber \\
& + \frac{5C_1 C_2^2 L_g^2}{12} \frac{1}{T}
 + \frac{K}{T},
\end{align}
where the first inequality holds due to the convexity of $F$, and the second equality holds since 
  $\mathbf x_t$ and $\mathbf y_t$ are implicit functions of i.i.d. random variables $\{ \mathbf w_k \}_{k=1}^{t-1}$ and $\{ \mathbf z_k \}_{k=1}^{t-1}$, and $\{ \mathbf w_t \}$ and $\{ \mathbf z_t \}$ are independent of each other. 

\subsection{Proof of Corollary\,2}
\label{sec: coro2}
Substituting $\eta_t = \frac{\alpha}{ \sigma  {t}}$ and 
$\beta_t = \frac{C_2}{{M(\mu)}t}$ into   RHS of \eqref{eq: appc_10}, we have
\begin{align}\label{eq: appe_1}
&\frac{1}{T} \sum_{t=2}^T \max\{  \frac{\alpha}{2\eta_t} - \frac{\alpha}{2\eta_{t-1}}  - \frac{\sigma}{2} ,0 \} R^2    =  0, \nonumber \\
&  \frac{s(m) L_1^2}{T}\sum_{t=1}^T \eta_t   \leq \frac{\alpha s(m) L_1^2 \log{T}}{\sigma T}  , \nonumber \\  
& \frac{M(\mu)^2 L_g^2}{4T}\sum_{t=1}^T \eta_t \beta_t^2
= \frac{ \alpha C_2^2 L_g^2}{4 \sigma T}  \sum_{t=1}^T \frac{1}{t^{3}} \leq 
\frac{ 3\alpha C_2^2 L_g^2}{8 \sigma T}, 
\end{align} 
where we have used the facts  that $\sum_{t=1}^T \frac{1}{t} \leq 1+ \log{T}  $ and \eqref{eq: appc_13}.
Based on \eqref{eq: appe_1} and  \eqref{eq: regret_ZOADMM_general}, we complete the proof.

\subsection{Proof of Corollary\,3}
\label{sec: coro3}

We consider the hybrid minibatch strategy
\begin{align}
\hat {\mathbf g}_t =\frac{1}{q_1 q_2}\sum_{j=1}^{q_1}  \sum_{i=1}^{q_2} 
\frac{f(\mathbf x_t + \beta_t \mathbf z_{t,j} ; \mathbf w_{t,i}) - f(\mathbf x_t  ; \mathbf w_{t,i}) }{\beta_t} \mathbf z_{t,j}  \label{eq: grad_mini_wz}
\end{align}
with
$
\hat {\mathbf g}_{t,ij} \Def \frac{f(\mathbf x_t + \beta_t \mathbf z_{t,j} ; \mathbf w_{t,i}) - f(\mathbf x_t   ; \mathbf w_{t,i}) }{\beta_t} \mathbf z_{t,j}$.
Based on  \eqref{eq: mu_gradient} and i.i.d. samples $\{ \mathbf w_{t,i} \}$ and  $\{ \mathbf z_{t,j} \}$, we have
\begin{align} \label{eq: def_gt_bar}
\bar{\mathbf g}_t \Def  \mathbb E [\hat {\mathbf g}_{t,ij}] =   \mathbb E[\mathbf g_t] +   \beta_t \mathbb E[L_{g,t} \nu(\mathbf x_t, \beta_t)] ,~ \forall i,j.
\end{align}
where for ease of notation we have replaced $L_g(\mathbf w_t)$ with $L_{g,t}$, 
$ \| \nu(\mathbf x_t, \beta_t)  \|_2 \leq \frac{1}{2} \mathbb E[ \| \mathbf z \|_2^3 ] \leq M(\mu) $ due to Assumption E.
From \eqref{eq: grad_mini_wz},  we obtain
\begin{align}
\mathbb E[ \| \hat {\mathbf g}_t \|_2^2 ]  =& \mathbb E \left [ \left  \| \frac{1}{q_1 q_2}\sum_{i=1}^{q_1} \sum_{j=1}^{q_2} (\hat {\mathbf g}_{t,ij}  - \bar{\mathbf g}_t) + \bar{\mathbf g}_t \right \|_2^2 \right ] \nonumber \\
=  & \| \bar{\mathbf g}_t \|_2^2 + \mathbb E \left [ \left  \| \frac{1}{q_1 q_2 }\sum_{i=1}^{q_1} \sum_{j=1}^{q_2}  (\hat {\mathbf g}_{t,ij}  - \bar{\mathbf g}_t) \right \|_2^2 \right ] \nonumber \\
 = &  \| \bar{\mathbf g}_t \|_2^2 + \frac{1}{q_1 q_2} 
 \mathbb E [ \left  \| \hat {\mathbf g}_{t,11}  - \bar{\mathbf g}_t \right \|_2^2  ] =\| \bar{\mathbf g}_t \|^2  \nonumber \\
 & + \frac{1}{q_1 q_2} \mathbb E[ \| \hat{\mathbf g}_{t,11} \|^2 ]  - \frac{1}{q_1 q_2}  \| \bar{\mathbf g}_{t} \|^2,
 \label{eq: appF_2}
\end{align}
where we have used the fact that $ \mathbb E [\hat {\mathbf g}_{t,ij}] =  \mathbb E [\hat {\mathbf g}_{t,11}]$ for any $i$ and $j$.

The definition of $\bar{\mathbf g}_t$ in \eqref{eq: def_gt_bar} yields
\begin{align}
\| \bar{\mathbf g}_t \|^2  \leq & 2 \| \mathbb E[\mathbf g_t]  \|_2^2 + 2 \| \beta_t \mathbb E[L_{g,t} \nu(\mathbf x_t, \beta_t)] \|_2^2 \nonumber \\
\leq & 2  \mathbb E[ \| \mathbf g_t \|_2^2 ]   + 2 \beta_t^2 \mathbb E[L_{g,t}^2] \mathbb E [ \|\nu(\mathbf x_t, \beta_t) \|_2^2 ] \nonumber \\
\leq &  2  \mathbb E[ \| \mathbf g_t \|_2^2 ] + \frac{1}{2} \beta_t^2 L_g^2 M(\mu)^2, \label{eq: appF_3}
\end{align}
where the first inequality holds due to Cauchy-Schwarz inequality, and the second inequality holds due to Jensen's inequality. 
From \eqref{eq: var_gradient}, we obtain
\begin{align}
\mathbb E[ \| \hat{\mathbf g}_{t,11} \|^2 ] \leq 2 s(m) \mathbb E [\| \mathbf g_t \|_2^2] + \frac{1}{2} \beta_t^2 L_g^2 M(\mu)^2. \label{eq: appF_4}
\end{align}
Substituting \eqref{eq: appF_3} and \eqref{eq: appF_4} into \eqref{eq: appF_2}, we obtain
\begin{align}
& \mathbb E[ \| \hat {\mathbf g}_t \|_2^2 ] \leq    \| \bar{\mathbf g}_t \|_2^2    +  \frac{1}{q_1 q_2} 
 \mathbb E [ \left  \| \hat {\mathbf g}_{t,11}   \right \|_2^2  ]  \nonumber \\
 \leq &  2 ( 1+\frac{ s(m)}{q_1 q_2} ) \mathbb E[ \| \mathbf g_t \|_2^2] + \frac{q_1 q_2+1}{2q_1 q_2} \beta_t^2
 L_g^2 M(\mu)^2. \label{eq: appF_5}
\end{align}

Similar to proof of Theorem\,1, substituting \eqref{eq: appF_5} into \eqref{eq: appc_7}, we obtain
\begin{align}\label{eq: appF_6}
& \overline{\mathrm{Regret}}_{T}(  \mathbf x_t, \mathbf y_t^\prime, \mathbf x^*, \mathbf y^* )  \nonumber \\
  \leq  &
\frac{1}{T} \sum_{t=2}^T \max\{  \frac{\alpha}{2\eta_t} - \frac{\alpha}{2\eta_{t-1}}  - \frac{\sigma}{2} , 0 \} R^2 \nonumber \\
&  +    \frac{1}{T}\sum_{t=1}^T \frac{\eta_t}{2} \mathbb E [ \|  \hat {\mathbf g}_t  \|_2^2 ]  + \frac{K}{T}\nonumber \\
\leq & \frac{1}{T} \sum_{t=2}^T \max\{  \frac{\alpha}{2\eta_t} - \frac{\alpha}{2\eta_{t-1}}  - \frac{\sigma}{2} , 0 \} R^2 \nonumber \\
& + \frac{( q_1 q_2 + s(m)  ) L_1^2}{q_1 q_2 T } \sum_{t=1}^T \eta_t \nonumber \\
& + \frac{(q_1 q_2+1)L_g^2 M(\mu)^2}{4q_1 q_2 T} \sum_{t=1}^T \eta_t \beta_t^2 + \frac{K}{T}.
\end{align}
Substituting
$\eta_t = \frac{C_1}{ \sqrt{1+\frac{s(m)}{q_1 q_2}} \sqrt{t}}$ and 
$\beta_t = \frac{C_2}{{M(\mu)}t}$ into \eqref{eq: appF_6}, we obtain
\begin{align}
&\overline{\mathrm{Regret}}_{T}(  \mathbf x_t, \mathbf y_t^\prime, \mathbf x^*, \mathbf y^* ) \nonumber \\
  \leq   &
\frac{\alpha R^2 }{2C_1} \frac{\sqrt{1+\frac{s(m)}{q_1 q_2}}}{\sqrt{T}} + 2 C_1 L_1^2 \frac{\sqrt{1+\frac{s(m)}{q_1 q_2}}}{\sqrt{T}} \nonumber \\
& + \frac{5C_1 C_2^2 L_g^2}{12T} \frac{q_1 q_2+1}{q_1 q_2 \sqrt{1+\frac{s(m)}{q_1 q_2}}} + \frac{K}{T} \nonumber \\
\leq & \frac{\alpha R^2 }{2C_1} \frac{\sqrt{1+\frac{s(m)}{q_1 q_2}}}{\sqrt{T}} + 2 C_1 L_1^2 \frac{\sqrt{1+\frac{s(m)}{q_1 q_2}}}{\sqrt{T}} \nonumber \\\
& +  \frac{5C_1 C_2^2 L_g^2}{6}  \frac{1}{T} + \frac{K}{T},
\label{eq: appF_7}
\end{align}
which then completes the proof.

\subsection{ZOO-ADMM for Sensor Selection}\label{sec: sel_sol}
We recall that the sensor selection problem can be cast as
\begin{align}
\begin{array}{ll}
\displaystyle \minimize_{\mathbf x,  \mathbf y} & \displaystyle \frac{1}{T} \sum_{t=1}^T f(\mathbf x; \mathbf w_t) + \mathcal I_1(\mathbf x) + \mathcal I_2(\mathbf y) \vspace*{0.05in} \\
\st &  \mathbf x - \mathbf y = \mathbf 0,
\end{array}
\label{eq: prob_sensorSel_online_ad}
\end{align}
where $\mathbf y \in \mathbb R^m$ is an auxiliary   variable,   $f(\mathbf x; \mathbf w_t) = - \mathrm{logdet}  (\sum_{i=1}^m  x_i \mathbf a_{i,t} \mathbf a_{i,t}^T  ) $ with $\mathbf w_t = \{\mathbf a_{i,t}\}_{i=1}^m$,  and  $\{ \mathcal I_i \}$ are indicator functions  
 \begin{align*}
\mathcal I_1(\mathbf x) = \left \{
\begin{array}{ll}
0 & \mathbf 0 \leq \mathbf x \leq \mathbf 1\\
\infty & \text{otherwise},
\end{array}
\right.
\mathcal I_2(\mathbf y) = \left \{
\begin{array}{ll}
0 & \mathbf 1^T \mathbf y = m_0\\
\infty & \text{otherwise}.
\end{array}
\right.
\end{align*}

Based on \eqref{eq: prob_sensorSel_online_ad},   two key steps of ZOO-ADMM \eqref{eq: x_step_ZOADMM}-\eqref{eq: y_step} are given by
 \begin{align}
&  \mathbf x_{t+1} = \argmin_{\mathbf 0 \leq \mathbf x \leq \mathbf 1} \left \{   
\left \| \mathbf x - \mathbf d_t  \right  \|_2^2 
 \right \} ,  \label{eq: x_step_sel} \\
&  \mathbf y_{t+1} = \argmin_{ \mathbf 1^T \mathbf y = m_0 }  \left  \{    \left \|  \mathbf y - \left (\mathbf x_{t+1} - (1/\rho) \boldsymbol \lambda_t  \right )
\right \|_2^2
  \right \},\label{eq: y_step_sel}  
\end{align}
where $\hat{\mathbf g}_t $ is the gradient estimate, and
$\mathbf d_t \Def 
 \frac{\eta_t}{\alpha} \left (
- \hat{\mathbf g}_t +  \boldsymbol \lambda_t - \rho  \mathbf x_t + \rho  \mathbf y_t  
\right )+ \mathbf x_t 
$.
Sub-problems \eqref{eq: x_step_sel} and \eqref{eq: y_step_sel}   yield closed-form solutions as below \citep{parikh2014proximal}
\begin{align}
 & [ \mathbf x_{t+1} ]_i =\left \{
\begin{array}{ll}
0 & \left [ \mathbf d_t \right ]_i  < 0 \\
\left [ \mathbf d_t \right ]_i  & \left [ \mathbf d_t \right ]_i  \in [0,1]\\
 1 & \left [ \mathbf d_t \right ]_i  > 1,
\end{array}
\right. \quad \text{and}   \\
&\mathbf y_{t+1} =  \mathbf x_{t+1}  - \frac{1}{\rho}\boldsymbol \lambda_t    + \frac{m_0 - \mathbf 1^T \left (\mathbf x_{t+1}  - \boldsymbol \lambda_t /\rho \right )}{m} \mathbf 1_m,
\end{align}
where $[\mathbf x]_i$ denote the $i$th entry of $\mathbf x$.

\subsection{ZOO-ADMM for Sparse Cox Regression}\label{sec: cox_sol}
This sparse regression problem can formulated as    
\begin{align}
\begin{array}{ll}
\displaystyle \minimize_{\mathbf x, \mathbf y} &  \displaystyle \frac{1}{n} \sum_{i=1}^n  f(\mathbf x;\mathbf w_i)  + \gamma \| {\mathbf y} \|_1 \\
\st & \mathbf x - \mathbf y = \mathbf 0,
\end{array}\label{eq: cox_reg_norm}
\end{align}
where  
$f(\mathbf x; \mathbf w_i) = \delta_i \left \{ - \mathbf a_i^T \mathbf x+  \log{(\sum_{j \in \mathcal R_i} e^{\mathbf a_j^T \mathbf x})}
\right \} 
$ with $\mathbf w_i = \mathbf a_i$.
By using  the ZOO-ADMM algorithm, we can avoid  the gradient calculation for the involved objective function in Cox regression.  The two key steps of ZOO-ADMM \eqref{eq: x_step_ZOADMM}-\eqref{eq: y_step} at iteration $i$ become
\begin{align}
& \mathbf x_{i+1} =   \frac{\eta_t}{\alpha} \left (
- \hat{\mathbf g}_i +   \boldsymbol \lambda_i - \rho  \mathbf x_i + \rho  \mathbf y_i  
\right )+ \mathbf x_i , \label{eq: x_step_cox} \\
& 
\mathbf y_{i+1} = \displaystyle \argmin_{  \mathbf y } \left \{  \| \mathbf y \|_1  + \frac{\rho}{2\gamma} \left \|  \mathbf y - \mathbf d_i
\right \|_2^2  \right \},
\label{eq: y_step_cox} 
\end{align}
where $\hat{\mathbf g}_i $ is the gradient estimate,   $\mathbf d_i = \left (\mathbf x_{i+1} - (1/\rho) \boldsymbol \lambda_i  \right )$, and the solution of sub-problem \eqref{eq: y_step_cox}  is given by the soft-thresholding operator at the point $\mathbf d_i$ with parameter $\rho/\gamma$ \citep[Sec.\,6]{parikh2014proximal} 
\begin{align*}
[\mathbf y_{i+1}]_k = \left \{
\begin{array}{ll}
(1-\frac{\gamma}{\rho  |[\mathbf d_i]_k|}) [\mathbf d_i]_k & [\mathbf d_i]_k > \frac{\gamma}{\rho } \\
0 & [\mathbf d_i]_k \leq \frac{\gamma}{\rho },
\end{array}
\right. 
\end{align*}
for $k = 1,2,\ldots, m$.

\bibliographystyle{abbrvnat}
\bibliography{Ref}

\end{document}